\documentclass[11pt]{article}
\usepackage{amsfonts}
\usepackage{multirow}
\usepackage{listings}
\usepackage[preprint]{acl}
\usepackage{booktabs} 
\usepackage{caption}  
\usepackage{times}
\usepackage{latexsym}
\usepackage{amsmath}
\usepackage{float}
\usepackage{hyperref}
\usepackage[T1]{fontenc}

\usepackage[utf8]{inputenc}

\usepackage{microtype}

\usepackage{inconsolata}

\usepackage{graphicx}

\title{Entropy Sentinel: Probing Entropy Traces for LLM Monitoring}

\author{Pedro Memoli Buffa \\
  Departamento de Matemática \\
  Universidad de Buenos Aires \\
  Buenos Aires, Argentina \\
  \texttt{pedromemolibuffa@uba.ar} \\
  \And
  Luciano Del Corro \\
  ELIAS Lab, Departamento de Ingenier\'ia \\ Universidad de San Andres \\
  Victoria, Argentina \\
  \texttt{delcorrol@udesa.edu.ar} \\}

\begin{document}

\maketitle






\begin{abstract}

Deploying LLMs raises two coupled challenges: (1) monitoring---estimating where a model underperforms as traffic drifts---and (2) prioritization---deciding where to intervene to close the largest performance gaps. We explore whether top-$k$ logprobs---cheap, consumer-accessible signals from standard inference---can serve as reliable proxies for domain-level quality of both verifiable and subjective tasks. We summarize each response's output-entropy profile into a compact vector, predict instance quality with a lightweight classifier, and then average predictions to yield a domain-level estimate. On verifiable tasks (ten STEM benchmarks, nine LLMs, exhaustive train/test compositions), estimates often track held-out accuracy, with several models showing near monotonic calibration. On subjective tasks, trained on LLM-judge scores over categorized real user conversations, several models track the judge's slice scores remarkably closely ($r$ up to $0.91$) and detect the worst-performing categories near-perfectly, though this is not the case for all LLMs. Where coupling holds, results suggest that entropy signals can support monitoring at a fraction of the cost of judge-based evaluation. Code and data \href{https://anonymous.4open.science/r/Entropy-Sentinel-E845/README.md}{here}.
\end{abstract}

\section{Introduction}
\label{sec:intro}

Deployed LLMs serve heterogeneous traffic that shifts over time. Yet practitioners still lack scalable answers to two tightly coupled questions: \emph{Where is the model underperforming on current usage?} and \emph{Where should we intervene to close those gaps?} In practice, these questions are addressed either with manually curated benchmarks—expensive, slow to update, and vulnerable to contamination—or with LLM-as-a-judge evaluation over production traces, which shows remarkable agreement with human judgment on chat conversations \cite{zheng2023judging}. While judge-based monitoring has become the de facto standard with frameworks like Langfuse \cite{langfuse} and Langsmith \cite{langsmith2023}, its cost scales linearly with traffic volume, restricting evaluation in practice to just a small portion of production data and leaving most traffic unmonitored.

\begin{figure}[h]
    \centering
    \includegraphics[width=0.8\linewidth]{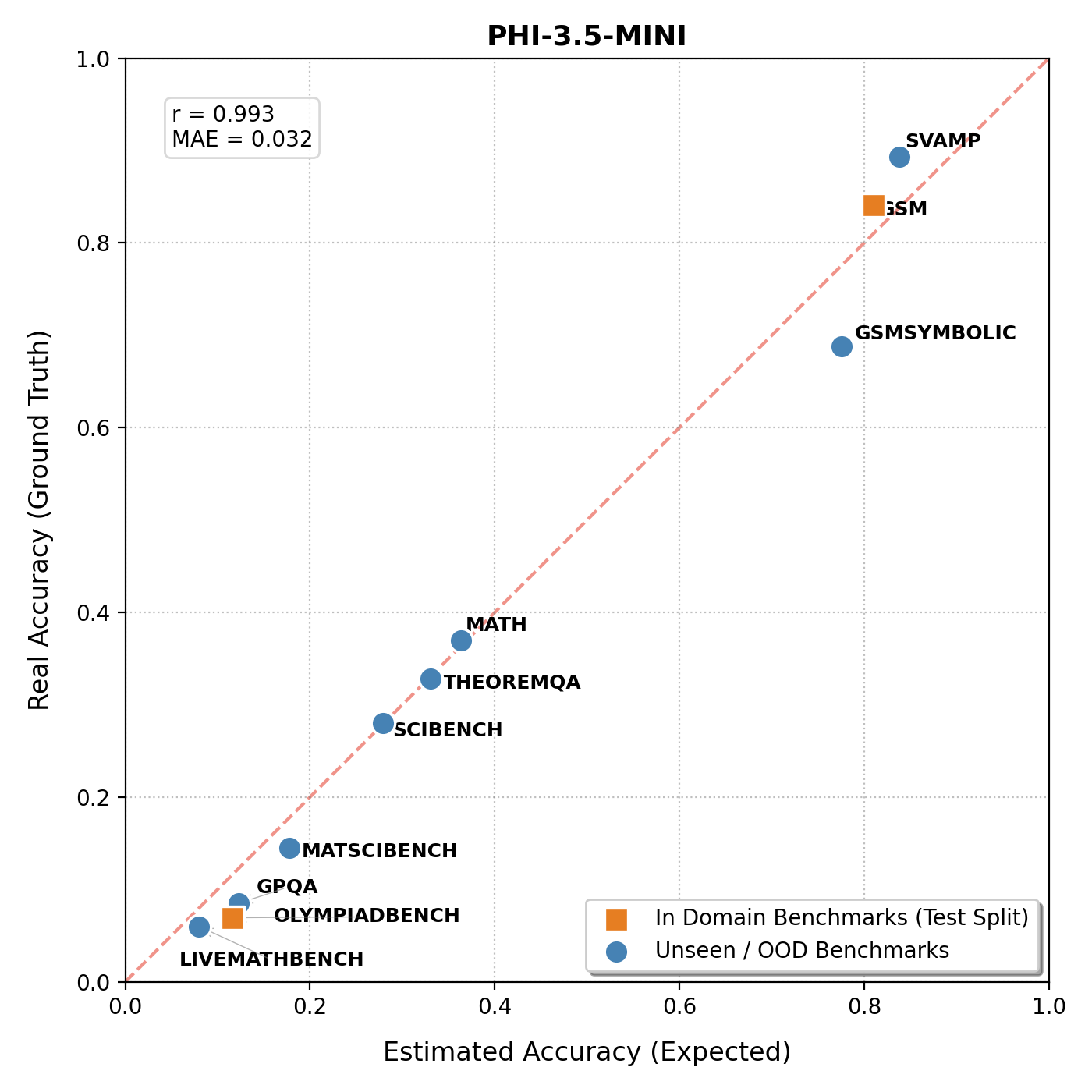}
    \caption{Entropy-based accuracy estimation for
    PHI-3.5-MINI-3.6B. Trained on two benchmarks (orange),
    the probe generalizes to eight unseen STEM benchmarks
    (blue).}
    \label{fig:phi3example}
\end{figure}

A natural alternative to scoring traces with an external judge is to probe response quality from cheap signals naturally produced during inference. If such an inexpensive trace could \emph{robustly} predict quality, we could then score slices of production traffic without repeated labeling or external large models; the resulting estimates immediately induce a ranking of slices, guiding downstream decisions such as intervention prioritization. For this to support monitoring at scale, the signal should be (i) cheap to extract, and (ii) available for both open and closed models, while the quality estimates must be (iii) robust to domain shift, and (iv) capable of aligning with judge scores in a realistic monitoring context.

Motivated by recent work showing that token-level entropy profiles form a signature for correctness, model family, and task type \cite{ali2025entropylens}, we pose the research question: \textbf{Can top-$k$ entropy signals support slice-level quality monitoring through a simple probe?} These signals satisfy (i) and (ii) by construction: they are produced naturally during decoding, are orders of magnitude smaller than hidden states (so their extraction does not compete for GPU bandwidth), and are exposed via top-$k$ log-probabilities by most model-serving APIs. To evaluate their capability for monitoring through a probe, we formalize \textbf{Entropy Sentinel} (ES) as the end-to-end estimation pipeline. For each response, ES summarizes the entropy trajectory into a compact feature vector and trains a lightweight model to predict instance-level quality against any per-instance label. These are binary correctness for verifiable tasks, or subjective judge scores for monitoring. Averaging predictions over a slice then yields a slice-level performance estimate. ES requires a single standard inference pass and no auxiliary LLMs.

We test (iii) by studying whether the probe can \emph{estimate out-of-distribution (OOD) accuracy on unlabeled slices} \cite{yu2024survey} for verifiable tasks. We evaluate on ten STEM QA benchmarks, whose verifiable correctness also enables an exhaustive study of ES's configuration space: for each $k\in\{1,2,3,4\}$ we train on all $\binom{10}{k}$ benchmark subsets and estimate accuracy on the remaining $10-k$, across nine LLMs (3B--20B, six families) and multiple estimator variants, totaling $>160{,}000$ configurations. Estimates often track held-out accuracy closely, with some LLMs presenting almost perfect calibration (Figure~\ref{fig:phi3example}), and supervision composition emerging as the dominant design factor.

To evaluate whether the probe can align with judge scores on OOD slices in a realistic monitoring setting (iv), we train ES on judge scores over a multi-turn chat benchmark grouped into 11 categories---thus emulating production traces clustered post hoc---and test whether it recovers the judge's category-level scores and rankings on held-out categories. Remarkably, several models track the judge's OOD category scores closely ($r$ up to $0.91$) and detect the worst-performing categories near-perfectly, with the signal not reducible to response length, though this coupling holds for only four of nine models, and strength in the verifiable setting does not always transfer. Where it does hold, results suggest that entropy signals can support monitoring and prioritization at a fraction of the cost of judge-based evaluation, validated on the target model and task distribution.

\section{From Signatures to Quality Estimates}
\label{sec:signatures}

To answer our research question, we formalize how Entropy Sentinel turns entropy trajectories into slice-level quality estimates through two stages: (1) summarize the decoding trace of each response into a compact uncertainty feature vector, and (2) train a lightweight probabilistic classifier to predict instance-level quality, whose predictions are aggregated into slice-level estimates.

\noindent\textbf{Entropy from top-$k$ log-probabilities.} For an input prompt $q$, let the model generate an output $\hat{y} = (y_1, \ldots, y_T)$. At decoding step $t$, let $p^{(t)}(\cdot)$ denote the next-token distribution conditioned on the prompt and previously generated tokens $(q, y_{<t})$. Many APIs expose only top-$k$ next-token probabilities at each step. We therefore approximate entropy by truncating the sum to the top-$k$ tokens: $\tilde{H}^{(t)} = -\sum_{i \in \mathrm{Top}\text{-}k} p^{(t)}_i \log p^{(t)}_i$, which differs from the true Shannon entropy because it omits the probability mass outside the Top-$k$ set.
We use $\tilde{H}^{(t)}$ as an uncertainty signal over the generated output tokens.

\noindent\textbf{From instance quality to slice estimates.}
Given a response $x=(q,\hat{y})$, we extract a feature vector from its entropy trajectory by summarizing $\{ \tilde{H}^{(t)} \}_{t=1}^{T}$ with a small set of statistics (Sec.~\ref{sec:features}), and train a probe that outputs an estimated measure of quality $\hat{P}(x)\in[0,1]$. For a domain (or slice) $D$ represented by a set of instances $X_D$, we estimate its quality by averaging the predictions:
\begin{equation}
\hat{A}(D) = \frac{1}{|X_D|}\sum_{x\in X_D}\hat{P}(x).
\label{eq:domain_acc}
\end{equation}
If $\hat{P}(x)$ is well-calibrated, then $\hat{A}(D)$ is a consistent estimator of the true domain measure. In the context of monitoring tasks with verifiable correctness, $\hat{P}$ is a probability of correctness and $\hat{A}$ is the accuracy of the slice. For monitoring LLM production traces, where there is no notion of objective ``correctness'', $\hat{P}$ can be trained on an external judge quality score normalized to $[0,1]$. Notably, the established ATC baseline for OOD accuracy estimation \cite{garg2022leveraging} can be viewed as a minimal instantiation of Eq.~\ref{eq:domain_acc}, with a hard threshold on a single score producing votes $\hat{P}(x)\in\{0,1\}$, applicable only when labels are binary. ES relaxes both restrictions, accepting multi-feature inputs and continuous labels, which makes this technique also suitable for monitoring.

\section{A Compact Entropy Signature}
\label{sec:features}

While entropy trajectories are known to carry quality signal \cite{ali2025entropylens}, they are variable-length sequences that cannot directly feed a simple probe. In this section we identify a compact, fixed-dimensional summary of the trajectory, by evaluating which summary statistics best discriminate correct from incorrect responses in STEM QA tasks with verifiable correctness.

\noindent\textbf{Setup.}
For each response, we compute the entropy trajectory $\{\tilde{H}^{(t)}\}_{t=1}^T$ from top-$k$ log-probabilities with $k{=}20$, matching the maximum exposed by commercial APIs (a full-vocabulary ablation in Appendix~\ref{sec:topk_ablation} shows little difference). Following standard evaluation practice in uncertainty quantification \citep{malinin2021uncertainty, kuhn2023semantic, bouchard2025uncertainty, kadavath2022language}, we score each \emph{single-number} summary statistic $s(\{\tilde{H}^{(t)}\})$ by its AUROC for discriminating incorrect responses, alongside traditional white-box UQ metrics (SEA, $\mathrm{NLL}_{\{\mathrm{avg,max,sum}\}}$, LNTP, MTP, PPL; as defined in Appendix~\ref{sec:baselinewhitebox}). Table~\ref{tab:entropy_analysis} reports two representative model--benchmark pairs at temperature $T{=}0.5$ \citep{kuhn2023semantic}; additional models and a temperature sensitivity analysis are in Appendices~\ref{sec:appendixentropyprofileresults} and~\ref{sec:appendix_sensibility_auroc} respectively.

\begin{table}[t]
\centering
\scriptsize

\setlength{\tabcolsep}{3pt}
\renewcommand{\arraystretch}{0.90}

\begingroup
\setlength{\aboverulesep}{0.2ex}
\setlength{\belowrulesep}{0.2ex}
\setlength{\lightrulewidth}{0.35pt}

\begin{tabular*}{\linewidth}{@{\extracolsep{\fill}} lccc @{}}
\toprule
\textbf{Statistic} & \textbf{MATH} & \textbf{GSM8K} & \textbf{OLYMP.} \\
\midrule
\multicolumn{4}{c}{\textit{\textsc{phi-3.5-mini 3.6B}}} \\
\midrule
Mean       & 0.7283          & 0.7391          & 0.7269          \\
Std Dev    & 0.7635          & 0.7397          & 0.7456          \\
Max        & 0.7952          & 0.7074          & 0.7778          \\
Q10        & \textbf{0.8248} & 0.7102          & 0.7589          \\
Q25        & 0.8084          & 0.7318          & 0.7323          \\
Q50        & 0.7357          & 0.7412          & 0.7028          \\
Q75        & 0.6939          & 0.7341          & 0.7027          \\
Q90        & 0.7130          & 0.7257          & 0.7280          \\
Skewness   & 0.5961          & 0.6902          & 0.6603          \\
Kurtosis   & 0.5726          & 0.6722          & 0.6497          \\
\midrule
SEA                     & \underline{0.8184} & \textbf{0.7649} & \textbf{0.7958} \\
$\text{NLL}_{\text{avg}}$ & 0.6983          & 0.7180          & 0.7188          \\
$\text{NLL}_{\text{max}}$ & 0.6937          & 0.6751          & 0.6768          \\
$\text{NLL}_{\text{sum}}$ & 0.8087          & \underline{0.7497} & \underline{0.7908} \\
LNTP                   & 0.6983          & 0.7180          & 0.7188          \\
MTP                    & 0.6937          & 0.6751          & 0.6768          \\
PPL                    & 0.6983          & 0.7180          & 0.7188          \\
\midrule
\multicolumn{4}{c}{\textit{\textsc{gpt-oss 20B}}} \\
\midrule
Mean       & 0.7736          & 0.7445          & 0.8583          \\
Std Dev    & 0.7771          & 0.7472          & 0.8498          \\
Max        & 0.8137 & 0.6991          & 0.7405          \\
Q10        & 0.7293          & 0.7243          & 0.8479          \\
Q25        & 0.7449          & 0.7177          & 0.8437          \\
Q50        & 0.7454          & 0.7233          & 0.8421          \\
Q75        & 0.7702          & 0.7476          & 0.8600          \\
Q90        & 0.7843 & 0.7483 & 0.8643 \\
Skewness   & 0.7676          & 0.7448          & 0.8558          \\
Kurtosis   & 0.7749          & 0.7506 & 0.8607 \\
\midrule
SEA                     & \textbf{0.9137} & \textbf{0.7886} & \textbf{0.8953} \\
$\text{NLL}_{\text{avg}}$ & 0.4127        & 0.3333          & 0.7641          \\
$\text{NLL}_{\text{max}}$ & 0.5351        & 0.5206          & 0.4042          \\
$\text{NLL}_{\text{sum}}$ & \underline{0.9070} & \underline{0.7864} & \underline{0.8869} \\
LNTP                   & 0.4127          & 0.3333          & 0.7641          \\
MTP                    & 0.5351          & 0.5206          & 0.4042          \\
PPL                    & 0.4127          & 0.3333          & 0.7641          \\
\bottomrule
\end{tabular*}
\endgroup

\vspace{-0.4em}
\caption{AUROC of entropy-profile summary stats across 3 benchmarks and 2 models (reporting $1-\text{AUROC}$ for skewness, kurtosis, LNTP, and MTP).}
\vspace{-0.8em}
\label{tab:entropy_analysis}
\end{table}

\begin{figure}[h]
    \centering
    \includegraphics[width=0.80\linewidth]{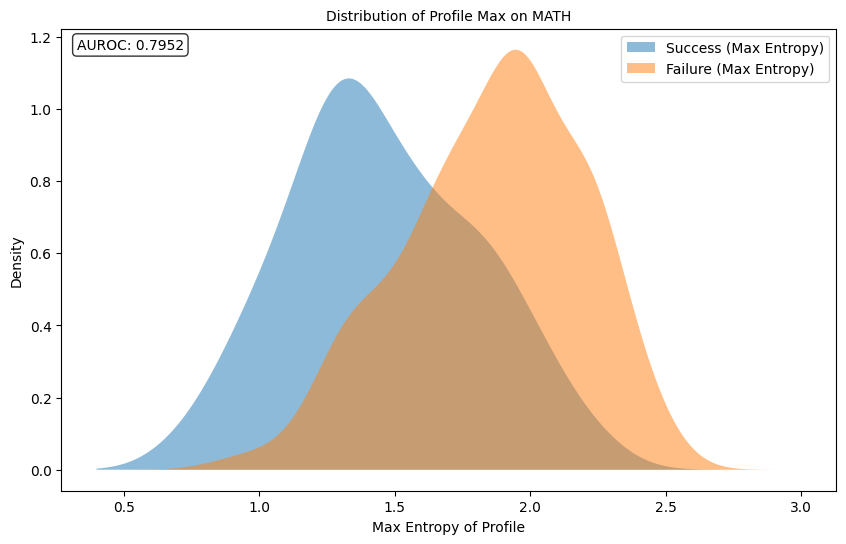}
    \caption{Max-entropy density for \textsc{phi-3.5-mini} on MATH (correct vs.\ incorrect). Incorrect responses shift to higher entropy, indicating greater uncertainty.}
        \label{fig:entropy_analysis}
\end{figure}

\noindent\textbf{No single summary is reliably best.}
The vast majority of statistics achieve AUROC $>0.5$, corresponding to correct responses concentrating in lower-entropy regions while incorrect ones shift higher (Figure~\ref{fig:entropy_analysis}). However, the top feature vary across models and benchmarks. Lower-tail quantiles (Q10/Q25) are most predictive on MATH, dispersion (Std) on GSM8K; accumulation metrics (SEA, $\text{NLL}_{\text{sum}}$) consistently rank among the strongest signals, while higher-order moments help in some settings but collapse toward chance in others. We consider then that committing to any single metric thus risks missing domain- and model-specific failure modes.

\noindent\textbf{A compact discriminative signature.}
We therefore encode each response with a fixed 17D entropy-profile vector capturing central tendency and dispersion (max, mean, std), distributional tails (Q10--Q90), shape (skewness, kurtosis), accumulation (SEA), and the white-box UQ metrics of Appendix~\ref{sec:baselinewhitebox}, truncating vocabulary sums to the top $k{=}20$ tokens where applicable. A lightweight classifier can then learn which aspect of entropy matters most for quality.

\section{Estimating OOD Accuracy on Verifiable STEM Tasks}
\label{sec:exp1}

We first evaluate whether the entropy signals support \emph{slice-level accuracy estimation under distribution shift} by training instance-level correctness predictors on a small set of labeled benchmarks, and estimate accuracy on unseen ones by aggregating predicted probabilities (Eq.~\ref{eq:domain_acc}). We scope this evaluation to STEM QA tasks with verifiable answers, whose unambiguous ground-truth labels also enable an exhaustive exploration of ES's configuration space in all its flexibility, thus establishing which design factors matter before we carry the conclusions into the monitoring experiment of Section~\ref{sec:exp2}.

\subsection{Experimental Setup}
\label{sec:exp1-setup}

\noindent\textbf{Benchmarks.}
We evaluate on ten STEM reasoning benchmarks spanning math and science:
GSM8K \cite{cobbe2021gsm8k}, SVAMP \cite{patel2021svamp}, GSM-Symbolic
\cite{mirzadeh2025gsmsymbolic}, MATH \cite{hendrycksmath2021}, TheoremQA
\cite{chen2023theoremqa}, SciBench \cite{wang2023scibench}, MatSciBench
\cite{zhang2025matscibench}, OlympiadBench \cite{he2024olympiadbench},
LiveMathBench \cite{livemathbench2025}, and GPQA \cite{rein2023gpqa}.
All tasks use zero-shot chain-of-thought prompting with free-form final
answers.

\noindent\textbf{Labels.}
An external validator LLM compares each model's extracted final answer
against the benchmark reference and outputs a binary correctness label
$z\in\{0,1\}$; a manual audit of 1000 instances yielded 99\% agreement
with human judgment (details in Appendix~\ref{sec:appendixsetup}). We
treat $z$ as supervision for the correctness predictor.

\noindent\textbf{Models.}
We run the full pipeline separately for nine LLMs from six families
(3B--20B): Ministral-3 (3B, 8B) \cite{ministral3}, Phi-3.5-Mini 3.8B
\cite{phi35mini}, Qwen-3 (4B, 8B) \cite{qwen3}, Gemma-3 (4B, 12B)
\cite{gemma3}, Llama-3.1 8B \cite{llama31}, and GPT-OSS 20B
\cite{gptoss}, restricting features to top-20 decoding logprobs.

\noindent\textbf{Train/test sweep.}
For each $k\in\{1,2,3,4\}$ and each benchmark subset $G$ of size $k$ ($\sum_{k=1}^{4}\binom{10}{k}=385$ groups), we train on 80\% of the instances from each benchmark in $G$ and estimate accuracy both on the held-out 20\% (in-domain) and on the remaining $10-k$ benchmarks, which are fully disjoint from the supervision set (OOD).

\noindent\textbf{Estimators.}
We explore \emph{flexible} and monitoring-friendly ES configurations by varying classifier family (logistic regression, random forest, MLP), class balancing, isotonic calibration, and four nested feature subsets (17D, 10D, 3D, 1D). Together with the nine models and 385 groups, this yields over 160{,}000 configurations (more details in Appendix~\ref{sec:appendixsetup}). We also evaluate the established ATC method for OOD accuracy estimation \cite{garg2022leveraging}, itself an instantiation of Eq.~\ref{eq:domain_acc} (Section~\ref{sec:signatures}), though restricted to binary labels and thus inapplicable to the monitoring setting of Section~\ref{sec:exp2}. Since ATC operates on a single scalar, we instantiate it separately for each individual feature (Section~\ref{sec:features}), yielding 17 ATC variants.

\noindent\textbf{Metrics.}
For each held-out benchmark we report \textbf{MAE}, the mean absolute error between estimated and true accuracy, as is standard in OOD slice accuracy estimation \cite{garg2022leveraging}. We also report Pearson $r$, since agreement metrics for monitoring are widely reported for monitoring \cite{zheng2023judging, li2024llmsjudges}.

\noindent\textbf{Implementation}
All LLMs are served with vLLM \cite{kwon2023efficient} (seed 42) with a maximum generation length of 2{,}048 tokens and temperature 0.5, as recommended for Shannon entropy estimation in \citet{kuhn2023semantic}. Classifiers are implemented in scikit-learn \cite{scikit-learn}, and features are z-scored using training-group statistics.

\begin{table}[t]
\centering
\scriptsize
\begin{tabular}{lcccc}
\toprule
& \multicolumn{2}{c}{\textbf{Extremes}} & \multicolumn{2}{c}{\textbf{Intermediate}} \\
\cmidrule(lr){2-3} \cmidrule(lr){4-5}
\textbf{Model} & \textbf{MAE} & \textbf{$r$} & \textbf{MAE} & \textbf{$r$} \\
\midrule
\textsc{Phi-3.5-Mini} (3.6B) & \textbf{0.03} & \textbf{0.99} & \textbf{0.09} & \textbf{0.99} \\
\midrule
\textsc{Ministral3} (3B) & 0.07 & 0.98 & 0.12 & 0.94 \\
\textsc{Ministral3} (8B) & 0.08 & 0.97 & 0.13 & 0.93 \\
\midrule
\textsc{Qwen3} (4B) & 0.08 & 0.93 & 0.13 & 0.98 \\
\textsc{Qwen3} (8B) & 0.13 & 0.90 & 0.17 & 0.84 \\
\midrule
\textsc{Gemma3} (4B) & 0.09 & 0.98 & 0.14 & 0.98 \\
\textsc{Gemma3} (12B) & 0.08 & 0.97 & 0.13 & 0.98 \\
\midrule
\textsc{Llama-3.1} (8B) & 0.07 & 0.98 & 0.13 & 0.98 \\
\midrule
\textsc{GPT-OSS} (20B) & 0.15 & 0.83 & 0.19 & 0.94 \\
\bottomrule
\end{tabular}
\caption{Cross-domain accuracy estimation with two \textit{a priori} training sets---\textit{Extremes} (GSM8K+OlympiadBench) and \textit{Intermediate} (MATH+SciBench)---using a calibrated, class-balanced random forest. $r$: Pearson correlation.}
\label{tab:apriori_results}
\end{table}


\subsection{Results}
\label{sec:exp1-results}

\noindent\textbf{Research Questions.} We organize results around four questions:
\textbf{(RQ1)} does a monitoring-compatible configuration support
OOD accuracy estimation?
\textbf{(RQ2)} how does it compare to ATC?
\textbf{(RQ3)} how sensitive is estimation quality to which benchmarks provide supervision?
\textbf{(RQ4)} how sensitive is it to estimator design choices
(classifier family, calibration, balancing, feature subset)?

\noindent\textbf{RQ1: Accuracy estimation under deployment-plausible defaults.}
We report results for two training groups chosen \textit{a priori} to reflect plausible supervision strategies rather than tuned for best score: \textbf{Extremes} (GSM8K + OlympiadBench), spanning the widest difficulty range, and \textbf{Intermediate} (MATH + SciBench), covering mid-range difficulty. Both use a random forest with class balancing, isotonic calibration, and the 10 entropy-distribution statistics, a configuration that could equally be trained as a regressor on judge scores. Table~\ref{tab:apriori_results} report estimation quality.

Both training groups generalize out of domain for most LLMs, but \textit{Extremes} is consistently stronger. Eight out of nine models achieve $r \ge 0.90$ with low error (MAE $0.03$--$0.13$), while \textit{Intermediate} yields systematically higher error (MAE $0.09$--$0.19$) despite comparable correlations. The signal is also model-dependent, \textsc{Phi-3.5-Mini} achieves near-perfect agreement (MAE $0.03$, $r = 0.99$), whereas \textsc{Qwen3-8B} couples weakest in both settings (MAE $0.13$--$0.17$, $r = 0.84$--$0.90$). 

We also found that for these configurations slice-level MAE is surprisingly decoupled from per-instance AUROC ($R^2 < 0.1$), suggesting the signal discriminates \emph{between slices} far more reliably than \emph{between instances}. This can be explained with entropy statistics being noisy but calibrated, so averaging over the slice cancels the per-instance noise that dominates correct--incorrect separation. Estimates stay accurate even where $\hat{P}(x)$ is a near-chance instance-level discriminator (Appendix~\ref{sec:comparison_MAE_auroc}).

\noindent\textbf{RQ2: How does it compare with ATC?}
While ATC is the established method for OOD accuracy estimation, its binary thresholding is unsuited to monitoring with continuous judge labels (Section~\ref{sec:exp2}), only the flexible configurations enable that setting. Here we measure what the flexibility trade-off is where both apply, comparing the RQ1 configuration against ATC on each individual feature in three different settings, since both require calibration data: (1) \emph{across all training compositions}, (2) \emph{Extremes} and (3) and \emph{Intermediate}.

Table~\ref{tab:es_vs_atc} reports the median MAE and IQR across all (composition, LLM) pairs. While top performing ATC variants achieve lower overall error and perform best under the \emph{Intermediate} regime, the flexible configuration ties for the lowest error on Extremes (.09 MAE). This parity, alongside model-specific gains where the flexible probe outperforms ATC (e.g., PHI’s MAE of 0.03 vs. ATC’s 0.06), demonstrates that non-linear models over multi-feature signatures can match or exceed the ATC baseline instantiation of ES. However, because higher-capacity models are more sensitive to training set selection and prone to overfitting under domain shift, realizing these gains requires more careful supervision design and hyperparameter tuning. The practical trade-off of flexibility is therefore the additional calibration effort required to outperform simpler baselines, with the benefit of being capable to support monitoring. For the next research questions we explore which factors most influence flexible instantiations of ES.

\newcommand{\miqr}[2]{#1\textsubscript{#2}}

\begin{table}[t]
\centering
\scriptsize
\setlength{\tabcolsep}{6pt}
\begin{tabular}{lccc}
\toprule
\textbf{Method} & \textbf{Overall} & \textbf{Extremes} & \textbf{Intermediate} \\
\midrule
ATC[max]                       & \textbf{\miqr{.08}{.03}} & \textbf{\miqr{.09}{.01}} & \miqr{.08}{.01} \\
ATC[$\text{NLL}_{\text{sum}}$] & \miqr{.09}{.03} & \miqr{.11}{.04} & \textbf{\miqr{.07}{.03}} \\
ATC[$\text{SE}_{\text{sum}}$]  & \miqr{.09}{.02} & \miqr{.12}{.04} & \miqr{.08}{.02} \\
ATC[Q10]                       & \miqr{.11}{.05} & \miqr{.11}{.02} & \miqr{.14}{.07} \\
ATC[Q25]                       & \miqr{.11}{.07} & \miqr{.11}{.05} & \miqr{.15}{.09} \\
ATC[Std]                       & \miqr{.12}{.09} & \miqr{.12}{.07} & \miqr{.12}{.10} \\
ATC[$\text{NLL}_{\text{max}}$] & \miqr{.13}{.14} & \miqr{.10}{.12} & \miqr{.10}{.11} \\
ATC[MTP]                       & \miqr{.13}{.14} & \miqr{.10}{.12} & \miqr{.10}{.11} \\
FLEXIBLE (RQ1 config)          & \miqr{.13}{.07} & \textbf{\miqr{.09}{.02}} & \miqr{.14}{.01} \\
ATC[Q50]                       & \miqr{.13}{.10} & \miqr{.13}{.06} & \miqr{.19}{.10} \\
ATC[Mean]                      & \miqr{.15}{.11} & \miqr{.14}{.09} & \miqr{.17}{.12} \\
ATC[Q90]                       & \miqr{.16}{.10} & \miqr{.16}{.09} & \miqr{.19}{.10} \\
ATC[Q75]                       & \miqr{.16}{.10} & \miqr{.16}{.07} & \miqr{.19}{.11} \\
ATC[$\text{NLL}_{\text{avg}}$] & \miqr{.17}{.06} & \miqr{.15}{.04} & \miqr{.22}{.07} \\
ATC[PPL]                       & \miqr{.17}{.06} & \miqr{.15}{.04} & \miqr{.22}{.07} \\
ATC[Skewness]                  & \miqr{.18}{.12} & \miqr{.17}{.09} & \miqr{.20}{.07} \\
ATC[Kurtosis]                  & \miqr{.19}{.13} & \miqr{.18}{.08} & \miqr{.21}{.06} \\
ATC[LNTP]                      & \miqr{.20}{.14} & \miqr{.24}{.11} & \miqr{.16}{.13} \\
\bottomrule
\end{tabular}
\caption{ES (RQ1 configuration) vs.\ per-feature ATC instantiations: median MAE over all 3,465 (training composition, LLM) pairs, reported overall and split by extremes vs.\ intermediate regimes. Subscripts give the IQR.}
\label{tab:es_vs_atc}
\end{table}

\noindent\textbf{RQ3: Sensitivity to training composition.}
Aggregating over all training groups of size $k\in\{1,2,3,4\}$ and flexible classifier configurations
(Table~\ref{tab:MAE_by_k}), increasing $k$ consistently lowers median MAE
(from $0.20$--$0.27$ at $k{=}1$ to $0.06$--$0.16$ at $k{=}4$) and sharply reduces
sensitivity to benchmark choice (IQR compressed to $0.03$--$0.04$), making training composition a first order factor in flexible ES performance for the evaluated task.

\begin{figure}[h]
    \centering
    \includegraphics[width=0.8\linewidth]{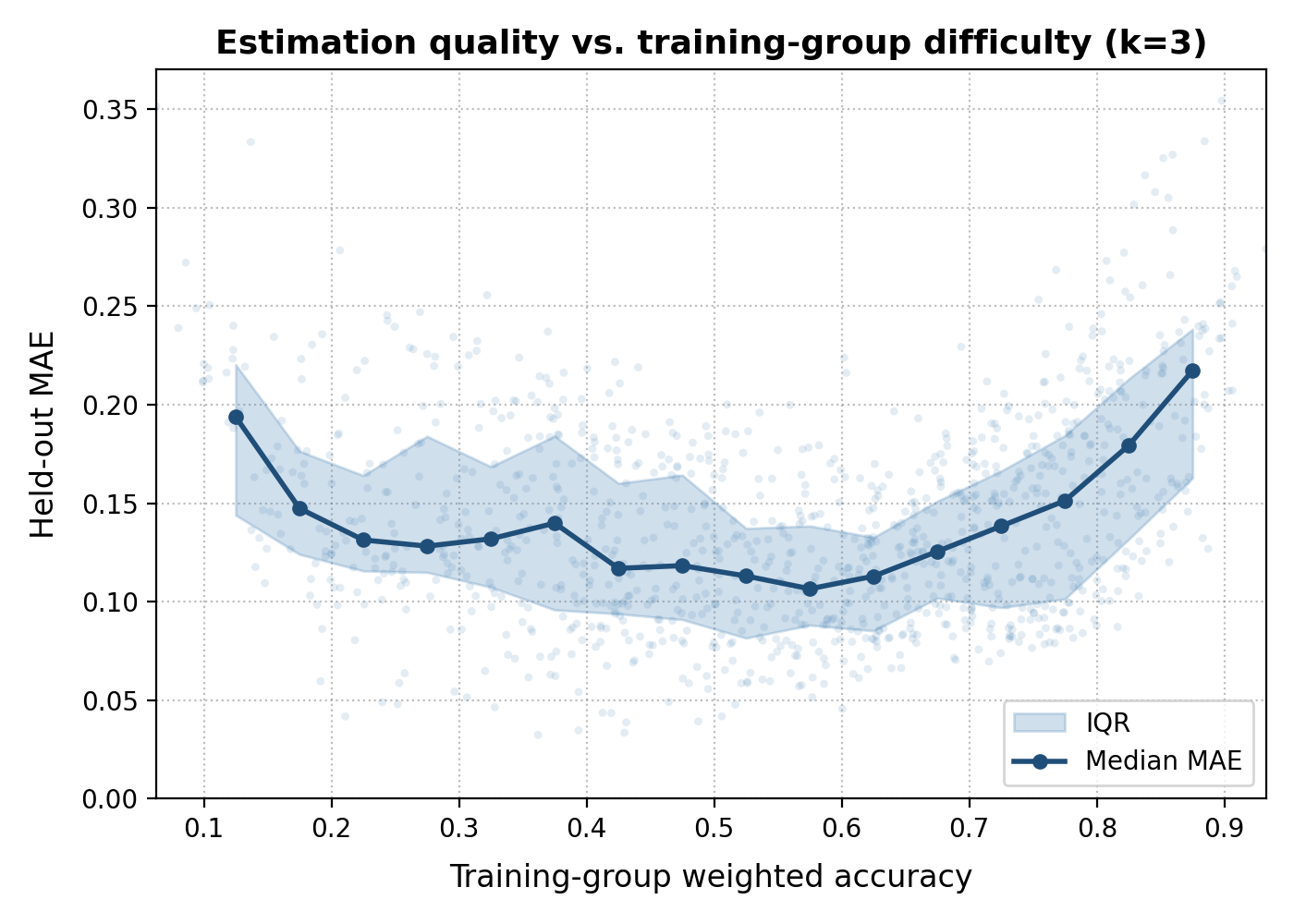}
    \caption{Held-out MAE vs.\ training-group weighted accuracy ($k{=}3$), aggregated
    across all nine LLMs. Estimation quality follows a U-shape: intermediate-difficulty
    groups minimize error, while very easy or very hard groups degrade generalization.
    Shaded region indicates IQR.}
    \label{fig:all_llm_accuracy}
\end{figure}

To interpret this effect, we summarize each training group by its
instance-weighted mean accuracy and relate it to held-out MAE under the RQ1 configuration as shown in Figure~\ref{fig:all_llm_accuracy}. The graph shows a clear U-shape, where groups that are too easy or too
hard generalize worse, while intermediate-difficulty groups ($\sim$$0.4$--$0.65$ weighted
accuracy) achieve the lowest error. We attribute this to difficulty-diverse supervision exposing the estimator to both success and failure patterns, whereas homogeneous difficulty groups miscalibrate under domain
shift. Best/worst combinations appear in
Appendix~\ref{sec:appendixsensibilityresults}.

\begin{table}
\centering
\scriptsize
\begin{tabular}{lcccc}
\toprule
& \multicolumn{4}{c}{\textbf{Training Benchmarks ($k$)}} \\
\cmidrule(lr){2-5}
\textbf{Model} & \textbf{1} & \textbf{2} & \textbf{3} & \textbf{4} \\
\midrule
\textsc{Phi-3.5-Mini} (3.6B) & .20\textsubscript{.11} & \textbf{.09}\textsubscript{.07} & \textbf{.07}\textsubscript{.04} & \textbf{.06}\textsubscript{.03} \\
\midrule
\textsc{Ministral3} (3B) & .24\textsubscript{.09} & .12\textsubscript{.07} & .10\textsubscript{.04} & .09\textsubscript{.03} \\
\textsc{Ministral3} (8B) & .20\textsubscript{.10} & .13\textsubscript{.07} & .11\textsubscript{.05} & .10\textsubscript{.04} \\
\midrule
\textsc{Qwen3} (4B) & .21\textsubscript{.08} & .12\textsubscript{.06} & .11\textsubscript{.04} & .09\textsubscript{.03} \\
\textsc{Qwen3} (8B) & .27\textsubscript{.06} & .21\textsubscript{.05} & .18\textsubscript{.04} & .16\textsubscript{.04} \\
\midrule
\textsc{Gemma3} (4B) & .26\textsubscript{.12} & .18\textsubscript{.08} & .13\textsubscript{.05} & .12\textsubscript{.04} \\
\textsc{Gemma3} (12B) & .23\textsubscript{.10} & .15\textsubscript{.07} & .12\textsubscript{.05} & .10\textsubscript{.04} \\
\midrule
\textsc{Llama-3.1} (8B) & .23\textsubscript{.09} & .14\textsubscript{.07} & .11\textsubscript{.05} & .11\textsubscript{.03} \\
\midrule
\textsc{GPT-OSS} (20B) & .22\textsubscript{.05} & .19\textsubscript{.06} & .17\textsubscript{.04} & .16\textsubscript{.04} \\
\bottomrule
\end{tabular}
\caption{Median MAE vs.\ $k$ benchmarks (IQR subscripts), aggregated over all groups and architectures.}
\label{tab:MAE_by_k}
\end{table}

\noindent\textbf{RQ4: Sensitivity to estimator design.}
We isolate estimator design effects by aggregating across all training groups and models; Table~\ref{tab:ablations} reports main effects on median MAE and Pearson $r$. Differences across all axes are small, with classifier and feature choice shifting median MAE by at most $0.02$--$0.03$, calibration and balancing having near-zero effect, and $r$ remaining at ${\approx}0.94$ throughout. Random forests and the full 17D profile perform best, while MLPs degrade slightly, likely from overfitting under limited supervision. Overall, estimator design is a second-order concern, the best and worst settings differ by ${\sim}0.035$ in median MAE, whereas training composition (RQ3) shifts it by up to $0.10$--$0.15$, the choice of supervision data thus dominates both feature dimensionality and classifier complexity. 

\begin{table}[t]
\centering
\scriptsize
\begin{tabular}{llcc}
\toprule
\textbf{Factor} & \textbf{Setting} & \textbf{MAE} & \textbf{$r$} \\
\midrule
\multirow{3}{*}{Classifier} & RF & \textbf{.11}\textsubscript{.07} & .94\textsubscript{.06} \\
& LR & .12\textsubscript{.06} & .94\textsubscript{.06} \\
& MLP & .15\textsubscript{.08} & .94\textsubscript{.06} \\
\midrule
Calibration & Y / N & .12\textsubscript{.07} / .13\textsubscript{.08} & .94\textsubscript{.06} / .94\textsubscript{.06} \\
Balancing & Y / N & .13\textsubscript{.07} / .13\textsubscript{.08} & .94\textsubscript{.06} / .94\textsubscript{.06} \\
\midrule
\multirow{4}{*}{Features} & 17D & \textbf{.11}\textsubscript{.06} & \textbf{.95}\textsubscript{.05} \\
& 10D & .14\textsubscript{.07} & .93\textsubscript{.07} \\
& 3D & .12\textsubscript{.07} & .94\textsubscript{.06} \\
& 1D & .14\textsubscript{.08} & .94\textsubscript{.06} \\
\bottomrule
\end{tabular}
\caption{Ablation main effects: median MAE and Pearson $r$ (IQR as subscripts), aggregated across all training groups and models. RF=Random Forest, LR=Logistic Regression, MLP=Multi Layer Perceptron.}
\label{tab:ablations}
\end{table}

\subsection{Takeaways}

Results provide very strong evidence that the entropy signals support OOD accuracy estimation, with the machinery on top of it being a second-order concern. The minimal ATC instantiation on a strong feature is the most accurate and stable probe on these binary tasks (RQ2), flexible configurations differ by at most ${\sim}0.035$ median MAE (RQ4), and the design factor that properly matters is supervision composition (RQ3), training groups that span difficulty consistently outperform homogeneous ones, which underrepresent either high-entropy failure or low-entropy success patterns. With just two well-chosen benchmarks, the probe often generalizes to eight unseen domains, in the best case (\textsc{Phi-3.5-Mini}) with near-perfect alignment to ground truth. 

\section{Monitoring LLM Traffic}
\label{sec:exp2}

Having established the top-k logprobs capability for OOD accuracy estimation on verifiable tasks, we now evaluate it in its intended setting of \emph{monitoring LLM traffic}, where responses are open-ended and no ground truth scores exists. We define quality in this setting by an LLM judge, and evaluate whether a simple probe can agree with its scores. Our hypothesis is that some LLMs, entropy signatures carry non-trivial quality signal even on ambiguous, non-verifiable tasks such as creative writing or open dialogue, where the correctness structure of the STEM setting is absent.

\subsection{Experimental Setup}
\label{sec:exp2-setup}

\noindent\textbf{Benchmark.}
We evaluate on WildBench \cite{lin2025wildbench}, which contains
challenging real user prompts from Chatbot Arena \cite{chiang2024chatbot} organized into 11 task categories. We cap conversations at two turns (1300 total turns) and set a maximum response length of 16{,}384 tokens.

\noindent\textbf{Labels.}
Each response is scored on a 1--10 scale by an external LLM judge
(GPT-5.4-Mini \cite{openai2026gpt54}) using the \texttt{single-v1} and
\texttt{single-v1-multi-turn} prompts from \citet{zheng2023judging};
\footnote{\url{https://github.com/lm-sys/FastChat/blob/main/fastchat/llm_judge/data/judge_prompts.jsonl}}
scores are normalized to $[0,1]$ and used as regression targets in
place of the binary labels of Section~\ref{sec:exp1-setup}.

\noindent\textbf{Models.}
We use the same nine LLMs as in the previous experiment (Sec~\ref{sec:exp1-setup}), served identically (vLLM, $T{=}0.5$, top-20 logprobs).

\noindent\textbf{Train/test sweep.}
Given the smaller per-category sample sizes, we adopt a leave-one-category-out (LOCO) protocol: for each of the 11 categories, we train on the remaining 10 and predict the held-out one, yielding 11 fully OOD category-level estimates. Because each estimate comes from a different training fold, we remove per-fold bias by centering each prediction on its fold's training-category mean, making results comparable \cite{enders2007centering}.

\noindent\textbf{Estimator.}
Since Section~\ref{sec:exp1} established that estimator design is second-order to supervision composition, we simply fix the configurations to a random forest regressor, with $n=200$ estimators and a depth of 8, balancing sample weight by category size.

\noindent\textbf{Metrics.}
We report two metrics, matching different goals of monitoring. For \emph{agreement}, we report \textbf{Pearson} $r$ between entropy-based and judge-assigned OOD category scores. For \emph{prioritization}, what matters is detecting the worst-performing categories, so we report the \textbf{AUROC} for identifying the bottom four categories under the judge from the entropy-based scores.

\subsection{Results}

We organize results into two research questions: \textbf{(RQ1)} Do entropy signals support OOD category agreement between estimated and real judge scores? \textbf{(RQ2)} Given that length biasing is a known confound of the judging protocol we use \cite{zheng2023judging}, how much of the result is attributed to the entropy distribution, and not just length?

\noindent\textbf{RQ1: OOD category monitoring with entropy signals.}
Table~\ref{tab:loco_results} reports our monitoring metrics for all nine LLMs. Overall, alignment between entropy-based and judge-assigned scores is \emph{strongly model-dependent}, more so than in Section~\ref{sec:exp1}. We observe that four models track the judge closely: \textsc{Phi-3.5-Mini}, \textsc{Llama-3.1-8B}, and both \textsc{Gemma-3} variants achieve $r = 0.79$--$0.91$ (all CIs excluding zero), with correspondingly strong detection of the worst categories. The remaining five models show weak or no linear coupling ($r = -0.22$--$0.54$, all CIs including zero), although bottom-category detection degrades more gracefully, with \textsc{Ministral3-8B} reaching an AUROC of $0.82$ despite its coupling not being significant, suggesting the signal may retain some prioritization value even where score-level agreement breaks down.

\begin{table}[h]
\centering
\scriptsize
\setlength{\tabcolsep}{5pt}
\begin{tabular}{lccc}
\toprule
\textbf{Model} & \textbf{$r$} & \textbf{95\% CI} & \textbf{AUROC}$_{\text{bot4}}$ \\
\midrule
\textsc{Phi-3.5-Mini} (3.6B) & \textbf{0.91} & $[0.68, 0.98]$ & 0.96$^{**}$ \\
\textsc{Llama-3.1} (8B)      & 0.89 & $[0.63, 0.97]$ & \textbf{1.00}$^{**}$ \\
\textsc{Gemma3} (12B)        & 0.84 & $[0.49, 0.96]$ & 0.82$^{\dagger}$ \\
\textsc{Gemma3} (4B)         & 0.79 & $[0.37, 0.94]$ & 0.82$^{\dagger}$ \\
\midrule
\textsc{GPT-OSS} (20B)       & 0.54 & $[-0.09, 0.86]$ & 0.75 \\
\textsc{Qwen3} (8B)          & 0.54 & $[-0.09, 0.86]$ & 0.71 \\
\textsc{Ministral3} (8B)     & 0.38 & $[-0.28, 0.80]$ & 0.82$^{\dagger}$ \\
\textsc{Qwen3} (4B)          & 0.22 & $[-0.44, 0.72]$ & 0.50 \\
\textsc{Ministral3} (3B)     & $-0.22$ & $[-0.72, 0.44]$ & 0.25 \\
\bottomrule
\end{tabular}
\caption{LOCO monitoring results per LLM, sorted by Pearson $r$ between entropy-based and judge-assigned category scores, with Fisher-$z$ 95\% confidence intervals. The midrule separates models whose CI excludes zero from the rest. AUROC significance is computed by exact enumeration over all $\binom{11}{4}=330$ bottom-four assignments: $^{**}$$p_{\text{perm}}<.01$, $^{\dagger}$$p_{\text{perm}}=.055$.}
\label{tab:loco_results}
\end{table}

Looking directly at each of the features individually (full table in the Appendix~\ref{sec:feature_monitoring_agreement}), we find that each of the strong models has several aspects of the entropy signature reaching $|r|>0.85$ with the category scores themselves, while the remaining weak models never go beyond $|r| \approx 0.67$. Interestingly, the strongest features are not the same as in Section~\ref{sec:exp1}. Accumulation metrics like SEA and $\text{NLL}_{\text{sum}}$ do very poorly here (just $r = 0.14$ for \textsc{Phi-3.5-Mini}), while simple averages like the mean and upper quantiles have the greatest $r$. We consider this result remarkable since it tells us that \emph{some models show substantial agreement with their uncertainty and very subjective judge scores on real chatbot traces}, suggesting that monitoring of similar Chatbot Arena-like traces is feasible for LLMs presenting this coupling.

\begin{figure}[h]
    \centering
    \includegraphics[width=0.8\linewidth]{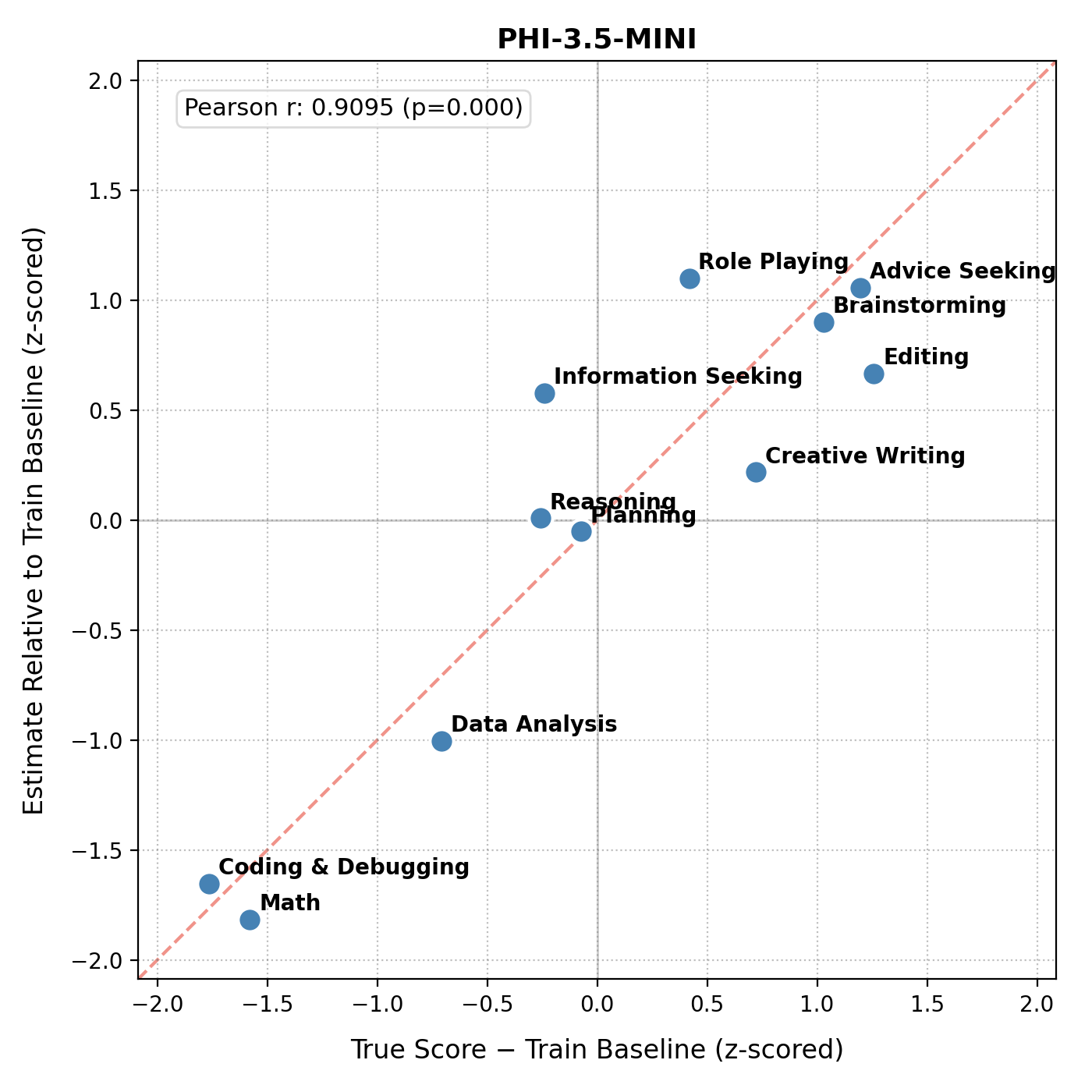}
    \caption{LOCO score estimation for \textsc{PHI-3.5} on WildBench ($r = 0.91$, 95\% CI $[0.68, 0.98]$). Each point is one held-out category, centered on its training fold's mean (z-scored).}
    \label{fig:placeholder}
\end{figure}

\noindent\textbf{RQ2: Studying the length confound.}
Since our strongest features (SEA, $\text{NLL}_{\text{sum}}$) accumulate over tokens (Section \ref{sec:features}), and our specific judge prompt is known to reward longer responses \cite{zheng2023judging}, we test whether our probe merely tracks response length or exploits genuine uncertainty. Under the identical estimator configuration, we compare ES against length alone (L) and ES augmented with length (ES+L) in Table~\ref{tab:length_ablation}. We observe that length alone is a markedly weaker predictor, reaching $r = 0.75$ only for \textsc{Llama-3.1-8B} and stays near or below zero for most models, never matching ES where ES is strong. More importantly, adding length to ES leaves results essentially unchanged (max $\Delta r = +0.05$, often negative), indicating that the entropy profile already subsumes any quality-relevant length information.
\begin{table}[h]
\centering
\scriptsize
\setlength{\tabcolsep}{5pt}
\begin{tabular}{lccc}
\toprule
& \multicolumn{3}{c}{\textbf{Pearson $r$}} \\
\cmidrule(lr){2-4}
\textbf{Model} & \textbf{L} & \textbf{ES} & \textbf{ES+L} \\
\midrule
\textsc{Phi-3.5-Mini} (3.6B) & 0.13 & \textbf{0.91} & \textbf{0.91} \\
\textsc{Llama-3.1} (8B)      & 0.75 & \textbf{0.89} & \textbf{0.89} \\
\textsc{Gemma3} (12B)        & 0.36 & 0.84 & \textbf{0.85} \\
\textsc{Gemma3} (4B)         & 0.37 & \textbf{0.79} & \textbf{0.79} \\
\textsc{GPT-OSS} (20B)       & $-0.09$ & 0.54 & \textbf{0.59} \\
\textsc{Qwen3} (8B)          & 0.40 & \textbf{0.54} & 0.51 \\
\textsc{Ministral3} (8B)     & $-0.14$ & \textbf{0.38} & 0.37 \\
\textsc{Qwen3} (4B)          & 0.00 & \textbf{0.22} & 0.13 \\
\textsc{Ministral3} (3B)     & \textbf{0.11} & $-0.22$ & $-0.13$ \\
\bottomrule
\end{tabular}
\caption{Length ablation under LOCO with the identical estimator configuration: Pearson $r$ for length alone (L), the entropy profile (ES), and both (ES+L). All L CIs intersect 0 except for \textsc{Llama-3.1 (8B)}}
\label{tab:length_ablation}
\end{table}

\subsection{Takeaways}

We found that entropy--quality coupling on subjective tasks is real but model-dependent. Four of nine models show significant coupling with the judge on OOD categories ($r = 0.79$--$0.91$, all 95\% CIs excluding zero), and two of them additionally separate the judge's worst categories at significance under an exact permutation test, with the signal not being reducible to response length (Table~\ref{tab:length_ablation}). While most LLMs show remarkable agreement in the verifiable setting of Section~\ref{sec:exp1}, this does not always transfer to subjective tasks, so entropy-based monitoring should be validated on both the target model and real traces clustered post hoc. Where the coupling holds, results suggest that entropy signals can support large-scale monitoring at a fraction of the cost of judge-based evaluation.

\section{Related Work}
\noindent\textbf{Estimating OOD performance.}
Estimating performance on unlabeled OOD data has been studied
extensively for classification networks \cite{garg2022leveraging, chen2021detecting}, grounded in the strong correlation
between in- and out-of-distribution accuracy \cite{miller2021accuracy}.
The established ATC baseline \cite{garg2022leveraging} thresholds a
single confidence score but is not adapted to LLM token sequences;
\citet{saxena2024predicting} extend ``agreement on the line''
\cite{baek2022agreement} to foundation models but require inference
from multiple models, which is prohibitive for continuous monitoring. We
instead use single-pass decoding traces, adapting ATC per-metric as
an in-method baseline (Section~\ref{sec:exp1-setup}).

\noindent\textbf{Uncertainty quantification.}
UQ methods for LLMs \cite{shorinwa2024survey, huang2023survey} sample
multiple generations \cite{kuhn2023semantic},
invoke external judges, probe hidden states \cite{azaria2023internal,
chen2024inside, zhang2025reasoning}, or aggregate token-level
uncertainties from logits \cite{malinin2021uncertainty,
fadeeva2024fact, bouchard2025uncertainty, moslonka2025learned,
vathul2025shed}; only the last meets our cost and API-access
requirements. Unlike this per-instance detection work, we target
continuous slice-level quality, an objective not necessarily aligned
with instance-level discrimination
(Appendix~\ref{sec:comparison_MAE_auroc}).

\noindent\textbf{Entropy-Lens.}
Entropy-Lens \cite{ali2025entropylens} classifies model family, task,
and correctness from entropy traces across transformer layers, but
requires residual-stream access; we use compact summaries of final-layer
top-$k$ logprobs alone.

\noindent\textbf{LLM-as-a-judge.}
LLM judges agree strongly with human preference \cite{zheng2023judging}
and underpin monitoring frameworks \cite{langfuse, langsmith2023},
but judging every trace scales with traffic; we ask whether a judge's
slice-level scores can be recovered from entropy profiles alone.

\section{Conclusion}
We studied whether top-$k$ entropy signals can support slice-level quality monitoring through a simple probe. On verifiable STEM tasks, entropy-based estimates often track held-out accuracy closely, with supervision composition as the dominant factor. On subjective monitoring, the coupling with judge scores is real but model-dependent, and strength in the verifiable setting does not guarantee transfer, so per-model validation remains necessary. Where the coupling holds, results suggest that cheap decoding signals can support monitoring to traffic that judge-based evaluation cannot affordably reach. Explaining which model properties drive the coupling remains future work.

\section{Limitations}
\label{sec:limitations}

\noindent\textbf{Unexplained model-dependence.}
Our most consequential finding is also our main open question: entropy--quality coupling varies strongly across models, and strength in the verifiable setting does not predict strength in the subjective one. We characterize this dependence but do not explain it, we didn't yet identify which architectural or training properties drive the coupling.

\noindent\textbf{Judge-relative supervision.}
In the monitoring experiment, quality is defined by a single LLM judge whose scores we treat as ground truth. Our claims are therefore about recovering the judge's assessments, not about quality itself; a probe that faithfully distills a biased judge inherits its biases. Whether the entropy signal tracks other judges, judge ensembles, or human preference directly remains untested.

\noindent\textbf{Sensitivity to decoding and formatting.}
Entropy traces depend on decoding choices (temperature, max length, stop criteria) and on prompting style. Changes in prompt instructions, answer formatting, or post-processing can shift entropy distributions without reflecting true capability changes, so a deployed sentinel should be retrained when the serving configuration changes.

\bibliography{custom}

@misc{ali2025entropylens,
  author = {Ali, Riccardo and Caso, Francesco and Irwin, Christopher and Liò, Pietro},
  title = {Entropy-Lens: The Information Signature of Transformer Computations},
  year = {2025},
  doi = {10.48550/arXiv.2502.16570},
  archiveprefix = {arXiv},
  journal = {arXiv.org},
}

@article{huang2023survey,
  author = {Huang, Lei and Yu, Weijiang and Ma, Weitao and Zhong, Weihong and Feng, Zhangyin and Wang, Haotian and Chen, Qianglong and Peng, Weihua and Feng, Xiaocheng and Qin, Bing and others},
  title = {A Survey on Hallucination in Large Language Models: Principles, Taxonomy, Challenges, and Open Questions},
  year = {2025},
  doi = {10.1145/3703155},
}

@article{shorinwa2024survey,
  author = {Shorinwa, Ola and Mei, Zhiting and Lidard, Justin and Ren, Allen Z. and Majumdar, Anirudha},
  title = {A Survey on Uncertainty Quantification of Large Language Models: Taxonomy, Open Research Challenges, and Future Directions},
  year = {2024},
  doi = {10.1145/3744238},
}

@article{kuhn2023semantic,
  author = {Kuhn, Lorenz and Gal, Yarin and Farquhar, Sebastian},
  title = {Semantic Uncertainty: Linguistic Invariances for Uncertainty Estimation in Natural Language Generation},
  journal = {International Conference on Learning Representations},
  year = {2023},
  doi = {10.48550/arXiv.2302.09664},
  eprint = {2302.09664},
  archiveprefix = {arXiv},
}

@article{malinin2021uncertainty,
  author = {Malinin, Andrey and Gales, Mark},
  title = {Uncertainty Estimation in Autoregressive Structured Prediction},
  year = {2021},
  doi = {10.17863/CAM.63497},
}

@article{fadeeva2024fact,
  author = {Fadeeva, Ekaterina and Rubashevskii, Aleksandr and Shelmanov, Artem and Petrakov, Sergey and Li, Haonan and Mubarak, Hamdy and Tsymbalov, Evgenii and Kuzmin, Gleb and Panchenko, Alexander and Baldwin, Timothy and others},
  title = {Fact-Checking the Output of Large Language Models via Token-Level Uncertainty Quantification},
  journal = {Findings of the Association for Computational Linguistics: ACL 2024},
  year = {2024},
  doi = {10.18653/v1/2024.findings-acl.558},
  publisher = {Association for Computational Linguistics},
}

@misc{kadavath2022language,
  author = {Saurav Kadavath and Tom Conerly and Amanda Askell and Tom Henighan and Dawn Drain and Ethan Perez and Nicholas Schiefer and Zac Hatfield-Dodds and Nova DasSarma and Eli Tran-Johnson and Scott Johnston and Sheer El-Showk and others},
  title = {Language Models (Mostly) Know What They Know},
  year = {2022},
  doi = {10.48550/arXiv.2207.05221},
  archiveprefix = {arXiv},
  journal = {arXiv.org},
}

@article{azaria2023internal,
  author = {Azaria, Amos and Mitchell, Tom},
  title = {The Internal State of an {LLM} Knows When It's Lying},
  journal = {Findings of the Association for Computational Linguistics: EMNLP 2023},
  year = {2023},
  doi = {10.18653/v1/2023.findings-emnlp.68},
  publisher = {Association for Computational Linguistics},
}

@article{chen2024inside,
  author = {Chen, Chao and Liu, Kai and Chen, Ze and Gu, Yi and Wu, Yue and Tao, Mingyuan and Fu, Zhihang and Ye, Jieping},
  title = {{INSIDE}: {LLMs}' Internal States Retain the Power of Hallucination Detection},
  journal = {International Conference on Learning Representations},
  year = {2024},
  doi = {10.48550/arXiv.2402.03744},
  eprint = {2402.03744},
  archiveprefix = {arXiv},
}

@article{moslonka2025learned,
  author = {Moslonka, Charles and Randrianarivo, Hicham and Garnier, Arthur and Malherbe, Emmanuel},
  title = {Learned Hallucination Detection in Black-Box {LLMs} using Token-level Entropy Production Rate},
  year = {2026},
  doi = {10.1007/978-3-032-21289-4_8},
  eprint = {2509.04492},
  archiveprefix = {arXiv},
}

@misc{vathul2025shed,
  author = {Vathul, Aneesh and Lee, Daniel and Chen, Sheryl and Tasmia, Arthi},
  title = {{ShED-HD}: A Shannon Entropy Distribution Framework for Lightweight Hallucination Detection on Edge Devices},
  year = {2025},
  doi = {10.48550/arXiv.2503.18242},
  archiveprefix = {arXiv},
  journal = {arXiv.org},
}

@article{bouchard2025uncertainty,
  author = {Bouchard, Dylan and Chauhan, Mohit Singh},
  title = {Uncertainty Quantification for Language Models: A Suite of Black-Box, White-Box, {LLM} Judge, and Ensemble Scorers},
  journal = {Transactions on Machine Learning Research},
  year = {2025},
  doi = {10.48550/arXiv.2504.19254},
  eprint = {2504.19254},
  archiveprefix = {arXiv},
}

@article{bouchard2025uqlm,
  author = {Bouchard, Dylan and Chauhan, Mohit Singh and Skarbrevik, David and Ra, Ho-Kyeong and Bajaj, Viren and Ahmad, Zeya},
  title = {{UQLM}: A {P}ython Package for Uncertainty Quantification in Large Language Models},
  journal = {Journal of Machine Learning Research},
  year = {2025},
  doi = {10.48550/arXiv.2507.06196},
  eprint = {2507.06196},
  archiveprefix = {arXiv},
}

@misc{zhang2025reasoning,
  author = {Zhang, Anqi and Chen, Yulin and Pan, Jane and Zhao, Chen and Panda, Aurojit and Li, Jinyang and He, He},
  title = {Reasoning Models Know When They're Right: Probing Hidden States for Self-Verification},
  year = {2025},
  doi = {10.48550/arXiv.2504.05419},
  archiveprefix = {arXiv},
  journal = {arXiv.org},
}

@misc{llama31,
Author = {Aaron Grattafiori and Abhimanyu Dubey and Abhinav Jauhri and Abhinav Pandey and Abhishek Kadian and Ahmad Al-Dahle and Aiesha Letman and Akhil Mathur and Alan Schelten and Alex Vaughan and Amy Yang and Angela Fan and Anirudh Goyal and Anthony Hartshorn and Aobo Yang and others},  title = {The Llama 3 Herd of Models},
  year = {2024},
  doi = {10.48550/arXiv.2407.21783},
  archiveprefix = {arXiv},
  journal = {arXiv},
}

@article{hendrycksmath2021,
  author = {Hendrycks, Dan and Burns, Collin and Kadavath, Saurav and Arora, Akul and Basart, Steven and Tang, Eric and Song, Dawn and Steinhardt, Jacob},
  title = {Measuring Mathematical Problem Solving With the {MATH} Dataset},
  journal = {NeurIPS},
  year = {2021},
  doi = {10.48550/arXiv.2103.03874},
  eprint = {2103.03874},
  archiveprefix = {arXiv},
}

@misc{cobbe2021gsm8k,
  author        = {Cobbe, Karl and Kosaraju, Vineet and Bavarian, Mohammad and Chen, Mark and Jun, Heewoo and Kaiser, Lukasz and Plappert, Matthias and Tworek, Jerry and Hilton, Jacob and others},
  title = {Training Verifiers to Solve Math Word Problems},
  year = {2021},
  doi = {10.48550/arXiv.2110.14168},
  archiveprefix = {arXiv},
  journal = {arXiv.org},
}

@article{patel2021svamp,
  author = {Arkil Patel and Satwik Bhattamishra and Navin Goyal},
  title = {Are {NLP} Models really able to Solve Simple Math Word Problems?},
  journal = {Proceedings of the 2021 Conference of the North American Chapter of the Association for Computational Linguistics: Human Language Technologies},
  year = {2021},
  doi = {10.18653/v1/2021.naacl-main.168},
  publisher = {Association for Computational Linguistics},
}

@article{mirzadeh2025gsmsymbolic,
author = {Iman Mirzadeh and Keivan Alizadeh and Hooman Shahrokhi and Oncel Tuzel and Samy Bengio and Mehrdad Farajtabar},
title = {{GSM-Symbolic}: Understanding the Limitations of Mathematical Reasoning in Large Language Models},
  journal = {International Conference on Learning Representations},
  year = {2024},
  doi = {10.48550/arXiv.2410.05229},
  eprint = {2410.05229},
  archiveprefix = {arXiv},
}

@article{rein2023gpqa,
  author = {Rein, David and Hou, Betty Li and Stickland, Asa Cooper and Petty, Jackson and Pang, Richard Yuanzhe and Dirani, Julien and Michael, Julian and Bowman, Samuel R.},
  title = {{GPQA}: A Graduate-Level Google-Proof {Q}\&{A} Benchmark},
  year = {2023},
  doi = {10.48550/arXiv.2311.12022},
  archiveprefix = {arXiv},
}

@misc{manakul2023selfcheckgpt,
Author = {Potsawee Manakul and Adian Liusie and Mark J. F. Gales},
Title = {SelfCheckGPT: Zero-Resource Black-Box Hallucination Detection for Generative Large Language Models},
Year = {2023},
Eprint = {arXiv:2303.08896},
}

@article{chen2023theoremqa,
  author    = {Chen, Wenhu and Yin, Ming and Ku, Max and Lu, Pan and Wan, Yixin and Ma, Xueguang and Xu, Jianyu and Wang, Xinyi and Xia, Tony},  title = {{TheoremQA}: A Theorem-driven Question Answering dataset},
  year = {2023},
  doi = {10.18653/v1/2023.emnlp-main.489},
  journal={Proceedings of the 2023 Conference on Empirical Methods in Natural Language Processing},
  publisher = {Association for Computational Linguistics},
}

@article{wang2023scibench,
  author = {Wang, Xiaoxuan and Hu, Ziniu and Lu, Pan and Zhu, Yanqiao and Zhang, Jieyu and Subramaniam, Satyen and Loomba, Arjun R. and Zhang, Shichang and Sun, Yizhou and Wang, Wei},
  title = {{SciBench}: Evaluating College-Level Scientific Problem-Solving Abilities of Large Language Models},
  year = {2024},
  doi = {10.48550/arXiv.2307.10635},
  eprint = {2307.10635},
  archiveprefix = {arXiv},
}

@article{he2024olympiadbench,
  author = {He, Chaoqun and Luo, Renjie and Bai, Yuzhuo and Hu, Shengding and Thai, Z. and Shen, Junhao and Hu, Jinyi and Han, Xu and Huang, Yujie and Zhang, Yuxiang and others},
  title = {{OlympiadBench}: A Challenging Benchmark for Promoting {AGI} with Olympiad-Level Bilingual Multimodal Scientific Problems},
  journal = {Proceedings of the 62nd Annual Meeting of the Association for Computational Linguistics (Volume 1: Long Papers)},
  year = {2024},
  doi = {10.18653/v1/2024.acl-long.211},
  publisher = {Association for Computational Linguistics},
}

@misc{zhang2025matscibench,
  author = {Zhang, Junkai and Gan, Jingru and Wang, Xiaoxuan and Jia, Zian and Gu, Changquan and Chen, Jianpeng and Zhu, Yanqiao and Ma, Mingyu Derek and Zhou, Dawei and Li, Ling and others},
  title = {{MatSciBench}: Benchmarking the Reasoning Ability of Large Language Models in Materials Science},
  year = {2025},
  doi = {10.48550/arXiv.2510.12171},
  archiveprefix = {arXiv},
  journal = {arXiv.org},
}

@article{kwon2023efficient,
  author = {Kwon, Woosuk and Li, Zhuohan and Zhuang, Siyuan and Sheng, Ying and Zheng, Lianmin and Yu, Cody Hao and Gonzalez, Joseph E. and Zhang, Hao and Stoica, Ion},
  title = {Efficient Memory Management for Large Language Model Serving with {PagedAttention}},
  year = {2023},
  doi = {10.1145/3600006.3613165},
}

@article{scikit-learn,
Author = {Fabian Pedregosa and Gaël Varoquaux and Alexandre Gramfort and Vincent Michel and Bertrand Thirion and Olivier Grisel and Mathieu Blondel and Andreas Müller and Joel Nothman and others},
Title = {Scikit-learn: Machine Learning in Python},
Year = {2012},
doi={10.48550/arXiv.1201.0490}
}

@article{miller2021accuracy,
Author = {John Miller and Rohan Taori and Aditi Raghunathan and Shiori Sagawa and Pang Wei Koh and Vaishaal Shankar and Percy Liang and Yair Carmon and Ludwig Schmidt},
title = {Accuracy on the Line: On the Strong Correlation Between Out-of-Distribution and In-Distribution Generalization},
  year = {2021},
  doi = {10.48550/arXiv.2107.04649},
  archiveprefix = {arXiv},
}

@article{enders2007centering,
  author = {Enders, Craig K. and Tofighi, Davood},
  title = {Centering Predictor Variables in Cross-Sectional Multilevel Models: A New Look at an Old Issue},
  journal = {Psychological Methods},
  year = {2007},
  doi = {10.1037/1082-989X.12.2.121},
  publisher = {American Psychological Association (APA)},
}

@misc{yu2024survey,
  author = {Yu, Han and Liu, Jiashuo and Zhang, Xingxuan and Wu, Jiayun and Cui, Peng},
  title = {A Survey on Evaluation of Out-of-Distribution Generalization},
  year = {2024},
  doi = {10.48550/arXiv.2403.01874},
  archiveprefix = {arXiv},
  journal = {arXiv.org},
}

@article{chiang2024chatbot,
  author = {Chiang, Wei-Lin and Zheng, Lianmin and Sheng, Ying and Angelopoulos, Anastasios Nikolas and Li, Tianle and Li, Dacheng and Zhang, Hao and Zhu, Banghua and Jordan, Michael I. and Gonzalez, Joseph E. and others},
  title = {Chatbot Arena: An Open Platform for Evaluating {LLMs} by Human Preference},
  year = {2024},
  doi = {10.48550/arXiv.2403.04132},
  eprint = {2403.04132},
  archiveprefix = {arXiv},
}

@inproceedings{garg2022leveraging,
  author    = {Garg, Saurabh and Balakrishnan, Sivaraman and Lipton, Zachary C. and Neyshabur, Behnam and Sedghi, Hanie},
  title     = {Leveraging Unlabeled Data to Predict Out-of-Distribution Performance},
  booktitle = {International Conference on Learning Representations},
  year      = {2022},
  eprint    = {2201.04234},
  archivePrefix = {arXiv},
  primaryClass  = {cs.LG},
  doi       = {10.48550/arXiv.2201.04234},
}

@article{chen2021detecting,
  author = {Chen, Jiefeng and Liu, Frederick and Avci, Besim and Wu, Xi and Liang, Yingyu and Jha, Somesh},
  title = {Detecting Errors and Estimating Accuracy on Unlabeled Data with Self-training Ensembles},
  journal = {NeurIPS},
  year = {2021},
  doi = {10.48550/arXiv.2106.15728},
  eprint = {2106.15728},
  archiveprefix = {arXiv},
}

@article{baek2022agreement,
  author = {Baek, Christina and Jiang, Yiding and Raghunathan, Aditi and Kolter, J. Zico},
  title = {Agreement-on-the-Line: Predicting the Performance of Neural Networks under Distribution Shift},
  journal = {Advances in Neural Information Processing Systems 35},
  doi = {10.52202/068431-1401},
  publisher = {Neural Information Processing Systems Foundation, Inc. (NeurIPS)},
}

@article{saxena2024predicting,
  author = {Rahul Saxena and Taeyoun Kim and Aman Mehra and Christina Baek and Zico Kolter and Aditi Raghunathan},
  title = {Predicting the Performance of Foundation Models via Agreement-on-the-Line},
  journal = {Advances in Neural Information Processing Systems 37},
  year = {2024},
  doi = {10.52202/079017-1002},
  publisher = {Neural Information Processing Systems Foundation, Inc. (NeurIPS)},
}

@misc{li2024llmsjudges,
  author = {Li, Haitao and Dong, Qian and Chen, Junjie and Su, Huixue and Zhou, Yujia and Ai, Qingyao and Ye, Ziyi and Liu, Yiqun},
  title = {{LLMs}-as-Judges: A Comprehensive Survey on {LLM}-based Evaluation Methods},
  year = {2024},
  doi = {10.48550/arXiv.2412.05579},
  archiveprefix = {arXiv},
  journal = {arXiv.org},
}

@article{zheng2023judging,
  author = {Zheng, Lianmin and Chiang, Wei-Lin and Sheng, Ying and Zhuang, Siyuan and Wu, Zhanghao and Zhuang, Yonghao and Lin, Zi and others},
  title = {Judging {LLM}-as-a-Judge with {MT}-Bench and Chatbot Arena},
  journal = {Advances in Neural Information Processing Systems 36},
  year = {2023},
  doi = {10.52202/075280-2020},
  eprint = {2306.05685},
  archiveprefix = {arXiv},
  publisher = {Neural Information Processing Systems Foundation, Inc. (NeurIPS)},
}

@article{lin2025wildbench,
  author = {Bill Yuchen Lin and Yuntian Deng and Khyathi Chandu and Faeze Brahman and Abhilasha Ravichander and Valentina Pyatkin and Nouha Dziri and Ronan Le Bras and Yejin Choi},
  title = {{WildBench}: Benchmarking {LLMs} with Challenging Tasks from Real Users in the Wild},
  year = {2025},
  doi = {10.48550/arXiv.2406.04770},
  eprint = {2406.04770},
  archiveprefix = {arXiv},
}

@misc{phi35mini,
Author = {Marah Abdin and Jyoti Aneja and Hany Awadalla and Ahmed Awadallah and Ammar Ahmad Awan and Nguyen Bach and Amit Bahree and Arash Bakhtiari and Jianmin Bao and others},
Title = {Phi-3 Technical Report: A Highly Capable Language Model Locally on Your Phone},
Year = {2024},
archiveprefix = {arXiv},
doi = {10.48550/arXiv.2404.14219},
journal = {arXiv},
}

@misc{qwen3,
  author = {Yang, An and Li, Anfeng and Yang, Baosong and Zhang, Beichen and Hui, Binyuan and Zheng, Bo and Yu, Bowen and Gao, Chang and Huang, Chengen and Lv, Chenxu and others},
  title = {Qwen3 Technical Report},
  year = {2025},
  doi = {10.48550/arXiv.2505.09388},
  archiveprefix = {arXiv},
  journal = {arXiv},
}

@misc{ministral3,
Author = {Alexander H. Liu and Kartik Khandelwal and Sandeep Subramanian and Victor Jouault and Abhinav Rastogi and Adrien Sadé and Alan Jeffares and Albert Jiang and Alexandre Cahill and others},
Title = {Ministral 3},
Year = {2026},
Eprint = {arXiv:2601.08584},
}

@misc{gemma3,
  author = {Kamath, Aishwarya and Ferret, Johan and Pathak, Shreya and Vieillard, Nino and Merhej, Ramona and Perrin, Sarah and Matejovicova, Tatiana and Ram'e, Alexandre and Rivière, Morgane and Rouillard, Louis and others},
  title = {Gemma 3 Technical Report},
  year = {2025},
  doi = {10.48550/arXiv.2503.19786},
  archiveprefix = {arXiv},
}

@misc{gptoss,
  author = {OpenAI},
  title = {gpt-oss-120b \& gpt-oss-20b Model Card},
  year = {2025},
  doi = {10.48550/arXiv.2508.10925},
  archiveprefix = {arXiv},
}

@misc{grok41fast,
  author = {{xAI}},
  title = {Grok 4.1 Fast},
  year = {2025},
  howpublished = {\url{https://data.x.ai/2025-11-17-grok-4-1-model-card.pdf}},
}

@article{livemathbench2025,
  author = {Liu, Junnan and Liu, Hongwei and Xiao, Linchen and Wang, Ziyi and Liu, Kuikun and Gao, Songyang and Zhang, Wenwei and Zhang, Songyang and Chen, Kai},
  title = {Are Your {LLMs} Capable of Stable Reasoning?},
  journal = {Findings of the Association for Computational Linguistics: ACL 2025},
  year = {2025},
  doi = {10.18653/v1/2025.findings-acl.905},
  eprint = {2412.13147},
  archiveprefix = {arXiv},
  publisher = {Association for Computational Linguistics},
}

@misc{openai2026gpt54,
  author = {{OpenAI}},
  title = {{ChatGPT} ({GPT-5.4} version)},
  year = {2026},
  howpublished = {\url{https://chatgpt.com}},
}

@misc{langsmith2023,
  author = {{LangChain, Inc.}},
  title = {{LangSmith}},
  year = {2023},
  howpublished = {\url{https://smith.langchain.com}},
}

@misc{langfuse,
  title = {Langfuse — Open‑Source AI Engineering Platform},
  author = {Rawert, Clemens and Klingen, Marc and Deichmann, Maximilian},
  year = {2023},
  note = {Software available from https://langfuse.com},
  url = {https://langfuse.com/}
}

\appendix

\section{Baseline White Box UQ Metrics}
\label{sec:baselinewhitebox}

We compare against nine standard uncertainty quantification baselines computed exclusively from the output token logprobs of a single forward pass. Most can be found in the uqlm library \cite{bouchard2025uqlm}. These constitute our most direct comparison: cheap signals available from any standard inference call, suitable for continuous performance monitoring.

\paragraph{Shannon Entropy (SE).} Following \citet{malinin2021uncertainty,manakul2023selfcheckgpt}, we compute token-level entropy over vocabulary $\mathcal{V}$:
\begin{equation}
    \text{SE}(y_i) = -\sum_{v \in \mathcal{V}} P(v \mid x, y_{<i}) \log P(v \mid x, y_{<i})
\end{equation}
In practice, the sum is truncated to the top $k=20$ tokens. We aggregate as: $\text{SE}_{\text{avg}}(y) = \frac{1}{L}\sum_{i=1}^{L} \text{SE}(y_i)$, $\text{SE}_{\text{max}}(y) = \max_{i} \text{SE}(y_i)$, and $\text{SE}_{\text{sum}}(y) = \sum_{i=1}^{L} \text{SE}(y_i)$ (Shannon Entropy Accumulation, abbreviated SEA).

\paragraph{Negative Log-Likelihood (NLL).} \citet{manakul2023selfcheckgpt} use $\text{NLL}(y_i) = -\log P(y_i \mid x, y_{<i})$, aggregated as: $\text{NLL}_{\text{avg}}(y) = \frac{1}{L}\sum_{i=1}^{L} \text{NLL}(y_i)$, $\text{NLL}_{\text{max}}(y) = \max_{i} \text{NLL}(y_i)$, and $\text{NLL}_{\text{sum}}(y) = \sum_{i=1}^{L} \text{NLL}(y_i)$.

\paragraph{Length-Normalized Token Probability (LNTP).} \citet{malinin2021uncertainty} propose the geometric mean of token probabilities:
\begin{equation}
    \text{LNTP}(y) = \left(\prod_{i=1}^{L} P(y_i \mid x, y_{<i})\right)^{1/L}
\end{equation}

\paragraph{Minimum Token Probability (MTP).} \citet{manakul2023selfcheckgpt} identify the weakest link: $\text{MTP}(y) = \min_{i} P(y_i \mid x, y_{<i})$.

\paragraph{Perplexity (PPL).} \citet{fadeeva2024fact} propose standard perplexity: $\text{PPL}(y) = \exp(\text{NLL}_{\text{avg}}(y))$.

\section{Effect of logprob truncation on predictive power}
\label{sec:topk_ablation}

Our entropy profiles use the top-20 logprobs from inference. To verify that this truncation does not distort predictive power, we
recomputed every Phi-3.5-mini profile from the \emph{full-vocabulary} log-softmax. The resulting AUROC deltas
(Table~\ref{tab:topk_ablation}) are negligible: the typical shift is
on the
order of $0.002$, and every location and scale statistic moves by
less than
$0.01$. The largest deltas concentrate in skewness and kurtosis on
the
harder benchmarks ($\approx 0.03$ on MATH and OlympiadBench,
$\approx 0.003$
on GSM8K), yet favor the $k=20$ truncation and remain small on the AUROC scale.

  \begin{table}[t]
  \centering
  \scriptsize
  \setlength{\tabcolsep}{4pt}
  \renewcommand{\arraystretch}{0.9}
  \begin{tabular}{lccc}
  \toprule
  \textbf{Stat} & \textbf{MATH} & \textbf{GSM8K} &
  \textbf{OLYMP.} \\
  \midrule
  Mean & $+0.002$ & $+0.000$ & $+0.002$ \\
  Std  & $+0.005$ & $-0.001$ & $+0.003$ \\
  Max  & $+0.007$ & $-0.000$ & $-0.007$ \\
  Q10  & $+0.000$ & $+0.001$ & $-0.000$ \\
  Q25  & $-0.001$ & $+0.002$ & $+0.002$ \\
  Q50  & $+0.004$ & $+0.002$ & $+0.000$ \\
  Q75  & $+0.002$ & $-0.001$ & $-0.001$ \\
  Q90  & $+0.004$ & $+0.002$ & $-0.000$ \\
  Skew & $-0.016$ & $-0.002$ & $-0.021$ \\
  Kurt & $-0.020$ & $-0.003$ & $-0.032$ \\
  SEA  & $+0.001$ & $-0.000$ & $+0.000$ \\
  \bottomrule
  \end{tabular}
  \caption{AUROC change (full vocabulary $-$ top-20) for Phi-3.5-Mini
  on MATH, GSM8K, and OlympiadBench.}
  \label{tab:topk_ablation}
  \end{table}

\section{Additional Entropy Profile Features Results}
\label{sec:appendixentropyprofileresults}

Table~\ref{tab:entropy_analysis_appendix} provides AUROC results for the remaining five models, supplementing the diagnostic study in Sec. 3. We observe significant heterogeneity across families and sizes. \textsc{Qwen-3 4B} achieves exceptionally high separability on MATH (Max: 0.9196), while its larger counterpart, \textsc{Qwen-3 8B}, shows a performance inversion with near-chance AUROCs on several statistics, even on equal benchmarks. This showcases that entropy profiles and their discriminative potential vary considerably between models of even the same family. Consistent with our main results, lower-tail quantiles (Q10, Q25) are most predictive for difficult reasoning tasks like MATH, while central tendency measures (Mean, Std Dev) perform better on elementary tasks like GSM8K. Furthermore, the \textsc{Gemma-3} family exhibits family-specific brittleness in higher-order moments (Skewness, Kurtosis), which collapse toward chance performance on difficult benchmarks.

Results provide supporting evidence that entropy signals carry a strong correctness signal across families, where the discriminatory potential of individual statistics is highly model and domain dependent.

\section{Feature Temperature Sensitivity Analysis}
\label{sec:appendix_sensibility_auroc}

We test whether the conclusions of Section~\ref{sec:features} are
sensitive to decoding choices. Holding model (Phi-3.5-Mini), benchmark
($500$ MATH-test problems), and item selection fixed, we run three seeds
at each of $T \in \{0.3, 0.5, 0.7, 1.0\}$; at $T{=}0.5$ the seed-$42$
run reuses the paper's original generations.
Table~\ref{tab:sensitivity_auroc} reports min--max AUROC ranges across
seeds.

We consider that results show that our conclusions are robust to temperature. Although individual statistics shift their AUROC --bulk and average-likelihood statistics (Mean, Std, NLL$_{\text{avg}}$, LNTP, PPL) have their AUROC drift monotonically upward with temperature-- the relative ordering of
features is preserved: accumulation metrics (SEA, NLL$_{\text{sum}}$)
and lower-tail quantiles (Q10, Q25) remain the strongest discriminators
at every temperature, staying within $0.78$--$0.84$ across all twenty
runs. Entropy Sentinel further marginalizes this per-instance variability by averaging over slices of hundreds or thousands of samples (Eq.~\ref{eq:domain_acc}).

\begin{table}[h]
\centering
\scriptsize
\setlength{\tabcolsep}{4pt}
\renewcommand{\arraystretch}{0.9}
\begin{tabular}{lcccc}
\toprule
\textbf{Stat} & $T=0.3$ & $T=0.5$ & $T=0.7$ & $T=1.0$ \\
\midrule
Mean              & .657--.702 & .657--.728 & .717--.746 & .724--.748 \\
Std               & .650--.694 & .668--.764 & .742--.762 & .752--.773 \\
Max               & .706--.717 & .723--.795 & .770--.790 & .767--.799 \\
Q10               & .812--.834 & .806--.830 & .829--.837 & .801--.831 \\
Q25               & .795--.820 & .794--.814 & .810--.823 & .783--.808 \\
Q50               & .715--.732 & .714--.744 & .750--.762 & .733--.736 \\
Q75               & .653--.693 & .659--.697 & .690--.719 & .697--.712 \\
Q90               & .664--.696 & .646--.713 & .694--.731 & .715--.745 \\
Skew              & .586--.647 & .582--.609 & .603--.647 & .603--.628 \\
Kurt              & .565--.628 & .557--.578 & .579--.618 & .579--.604 \\
SEA               & .787--.816 & .805--.826 & .803--.839 & .812--.815 \\
NLL$_{\text{avg}}$ & .629--.660 & .625--.698 & .688--.716 & .713--.737 \\
NLL$_{\text{max}}$ & .664--.693 & .669--.698 & .671--.708 & .679--.705 \\
NLL$_{\text{sum}}$ & .769--.787 & .774--.817 & .785--.830 & .798--.813 \\
LNTP              & .629--.660 & .625--.698 & .688--.716 & .713--.737 \\
MTP               & .664--.693 & .669--.698 & .671--.708 & .679--.705 \\
PPL               & .629--.660 & .625--.698 & .688--.716 & .713--.737 \\
\bottomrule
\end{tabular}
\caption{Per-statistic AUROC ranges (min--max across seeds $s \in \{42, 43,
44\}$) for Phi-3.5-Mini on $500$ MATH-test problems at four sampling
temperatures. At $T = 0.5$ the seed-$42$ run uses the paper's original
generations rather than a regenerated sample. Skew, Kurt, LNTP, and MTP
report $1 - \text{AUROC}$, following the convention of
Table~\ref{tab:entropy_analysis}.}
\label{tab:sensitivity_auroc}
\end{table}

\section{Experimental Setup Details for STEM Evaluation}
\label{sec:appendixsetup}

This appendix collects the setup details omitted from
Section~\ref{sec:exp1-setup}.

\subsection{Benchmark Preprocessing}
All tasks use zero-shot prompting with free-form final
answers. For benchmarks originally multiple-choice (GPQA, SciBench), we remove answer options from the prompt. For OlympiadBench we restrict to text-only math and physics questions; for LiveMathBench we use the
\texttt{v202505\_all\_en}
subset.\footnote{\url{https://huggingface.co/datasets/opencompass/LiveMathBench}}
When a split exists, we evaluate on the test portion; otherwise we
evaluate on the full benchmark.

\subsection{Instance Labeling}
Each benchmark instance provides a question $q$, a reference answer
$y^\star$, and a model-generated response $\hat{y}$. We extract the
model's final answer from $\hat{y}$ using benchmark-specific
post-processing (e.g., stripping formatting and selecting the last
boxed/numeric expression when applicable). An external \emph{validator}
LLM (\textsc{Grok-4.1-Fast-Reasoning} \cite{grok41fast}) receives
$(q,\hat{y}_{\text{final}},y^\star)$ and outputs a binary label
$z\in\{0,1\}$ indicating whether the final answer matches the reference.
A manual audit of 1000 randomly sampled instances yielded $99\%$
agreement with human judgment. 

\subsection{Evaluation and Verification Prompt}

Ground-truth correctness labels were produced using \textsc{Grok-4.1-Fast-Reasoning} \cite{grok41fast} via the LiteLLM API as an external validator. The validator receives the original question, the model's response, and the reference ground-truth answer, then outputs a structured binary decision using a Pydantic response schema (\texttt{class Response(BaseModel): success: bool}).

The system prompt instructs the validator to determine answer equivalence under benchmark-specific criteria (e.g., symbolic or numeric equivalence):

\begin{quote}
\small\ttfamily
\textbf{System:} Task: Determine if the Response corresponds to the Correct Answer for the Question, based on the given Correct Answer text. Answer ONLY with the exact format: \{"success": True\} or \{"success": False\}
\end{quote}

\noindent The user prompt provides the evaluation context:

\begin{quote}
\small\ttfamily
\textbf{User:}
\# Question\\
\{question\}\\[0.5em]
\# Response\\
\{cleaned\_response\}\\[0.5em]
\# Correct Answer\\
\{correct\_answer\}
\end{quote}

\noindent Responses are preprocessed by removing special tokens (e.g., \texttt{<|end|>}, \texttt{<|endoftext|>}) before validation. The structured Pydantic output ensures consistent binary classification across all 385 training configurations and benchmark evaluations.

\subsection{Feature Interface}
We restrict features to top-20 decoding logprobs to match a common
interface constraint in logprob-returning serving stacks; see
Appendix~\ref{sec:topk_ablation} for an ablation against the
full-vocabulary distribution.

\subsection{Estimator Configurations}
We evaluate three classifier families: logistic regression, random
forest, and a multilayer perceptron (MLP). For each classifier, we
select hyperparameters via a 5-fold cross-validated grid search on the
training group, optimizing ROC-AUC. Logistic regression uses no
hyperparameter search. For random forests (100 estimators), we search
max depth $\in \{3, 5, 10\}$ and min samples split $\in \{2, 5, 10\}$.
For MLPs (ReLU activations, early stopping, $\alpha = 0.001$), we
search hidden layer sizes $\in
\{(5),(8),(10),(15),(20),(8,4),(10,5),(15,8)\}$.

We also vary two training choices: isotonic calibration (on/off) and
class balancing (on/off), the latter implemented via scikit-learn's
built-in mechanisms. Logistic regression uses
\texttt{class\_weight="balanced"}, reweighting the loss inversely
proportional to class frequencies; random forests use
\texttt{class\_weight="balanced\_subsample"}, reweighting within each
bootstrap sample. For MLPs, which lack native class weighting, we apply
random oversampling via \texttt{RandomOverSampler} from
imbalanced-learn\footnote{\url{https://imbalanced-learn.org/stable/}}
prior to training.

The four nested feature subsets are: (i) the full 17D profile, (ii) the
10 entropy-distribution statistics only, (iii) max,
$\text{SE}_{\text{sum}}$, and $\text{NLL}_{\text{sum}}$---the three most
consistently discriminative signals from the AUROC analysis in
Section~\ref{sec:features}---and (iv) $\text{SE}_{\text{sum}}$ alone,
the single strongest individual metric. Across 9 models, 385 train/test
groups, 3 classifier families, 2$\times$2 training options, 4 feature
subsets, and the calibrated single-metric baselines, this produces over
160{,}000 configurations.

\begin{table*}[t]
  \centering
  \scriptsize
  \setlength{\tabcolsep}{3pt}
  \renewcommand{\arraystretch}{0.88}
  
  \begin{tabular*}{\textwidth}{@{\extracolsep{\fill}} lccc @{\hspace{1.5em}} lccc @{}}
  \toprule
  \textbf{Statistic} & \textbf{MATH} & \textbf{GSM8K} & \textbf{OLYMP.} & \textbf{Statistic} & \textbf{MATH} & \textbf{GSM8K} & \textbf{OLYMP.} \\
  \midrule
  \multicolumn{4}{c}{\textit{\textsc{Ministral-3 3B}}} & \multicolumn{4}{c}{\textit{\textsc{Qwen-3 8B}}} \\
  \midrule
  Mean       & 0.7512 & 0.7823 & 0.8225 & Mean       & 0.5698 & 0.7726 & 0.7935 \\
  Std Dev    & 0.7648 & 0.7873 & 0.8077 & Std Dev    & 0.5966 & 0.7776 & 0.8104 \\
  Max        & 0.7856 & 0.7771 & 0.7764 & Max        & 0.5392 & 0.7117 & 0.7323 \\
  Q10        & 0.7887 & 0.7741 & 0.8099 & Q10        & \textbf{0.6484} & 0.7509 & \textbf{0.8236} \\
  Q25        & 0.7749 & 0.7679 & 0.8230 & Q25        & 0.5926 & 0.7614 & \underline{0.8128} \\
  Q50        & 0.7558 & 0.7726 & 0.8328 & Q50        & 0.5279 & 0.7485 & 0.7820 \\
  Q75        & 0.7332 & 0.7775 & 0.8259 & Q75        & 0.5593 & 0.7701 & 0.7829 \\
  Q90        & 0.7504 & 0.7754 & 0.8130 & Q90        & 0.5914 & 0.7802 & 0.8041 \\
  Skewness   & 0.7145 & 0.7457 & 0.8229 & Skewness   & 0.5520 & 0.7633 & 0.7699 \\
  Kurtosis   & 0.7113 & 0.7368 & 0.8199 & Kurtosis   & 0.5617 & 0.7652 & 0.7698 \\
  \midrule
  SEA        & \textbf{0.8612} & \textbf{0.8145} & \textbf{0.8700} & SEA        & \underline{0.6014} & \textbf{0.8032} & 0.7950 \\
  $\text{NLL}_{\text{avg}}$ & 0.6168 & 0.7256 & 0.7864 & $\text{NLL}_{\text{avg}}$ & 0.5689 & 0.7701 & 0.7907 \\
  $\text{NLL}_{\text{max}}$ & 0.3681 & 0.4432 & 0.4345 & $\text{NLL}_{\text{max}}$ & 0.5125 & 0.6587 & 0.6028 \\
  $\text{NLL}_{\text{sum}}$ & \underline{0.8571} & \underline{0.8009} & \underline{0.8661} & $\text{NLL}_{\text{sum}}$ & 0.5993 & \underline{0.8012} & 0.7923 \\
  LNTP       & 0.6168 & 0.7256 & 0.7864 & LNTP       & 0.5689 & 0.7701 & 0.7907 \\
  MTP        & 0.3681 & 0.4432 & 0.4345 & MTP        & 0.5125 & 0.6587 & 0.6028 \\
  PPL        & 0.6168 & 0.7256 & 0.7864 & PPL        & 0.5689 & 0.7701 & 0.7907 \\
  \midrule
  \multicolumn{4}{c}{\textit{\textsc{Qwen-3 4B}}} & \multicolumn{4}{c}{\textit{\textsc{Llama-3 8B}}} \\
  \midrule
  Mean       & 0.8759 & 0.8093 & 0.8679 & Mean       & 0.5557 & 0.7319 & 0.5263 \\
  Std Dev    & 0.9011 & 0.8100 & 0.8849 & Std Dev    & 0.6431 & 0.7602 & 0.5875 \\
  Max        & 0.9196 & 0.8144 & 0.8472 & Max        & 0.8404 & 0.7520 & 0.7465 \\
  Q10        & 0.8607 & 0.7452 & 0.8705 & Q10        & 0.6763 & 0.7542 & 0.5753 \\
  Q25        & 0.8373 & 0.7605 & 0.8631 & Q25        & 0.6614 & 0.7538 & 0.5861 \\
  Q50        & 0.8133 & 0.7937 & 0.8523 & Q50        & 0.5813 & 0.7097 & 0.5372 \\
  Q75        & 0.8299 & \underline{0.8148} & 0.8571 & Q75        & 0.5091 & 0.7110 & 0.4946 \\
  Q90        & 0.8712 & 0.8003 & 0.8732 & Q90        & 0.5784 & 0.7470 & 0.5392 \\
  Skewness   & 0.8185 & 0.7649 & 0.8466 & Skewness   & 0.4448 & 0.6457 & 0.4539 \\
  Kurtosis   & 0.8153 & 0.7485 & 0.8483 & Kurtosis   & 0.4494 & 0.6378 & 0.4578 \\
  \midrule
  SEA        & \textbf{0.9327} & \textbf{0.8193} & \underline{0.8979} & SEA        & \textbf{0.8861} & \textbf{0.7877} & \textbf{0.8555} \\
  $\text{NLL}_{\text{avg}}$ & 0.8746 & 0.7869 & 0.8683 & $\text{NLL}_{\text{avg}}$ & 0.5400 & 0.7192 & 0.5127 \\
  $\text{NLL}_{\text{max}}$ & 0.8526 & 0.7348 & 0.7899 & $\text{NLL}_{\text{max}}$ & 0.7957 & 0.6758 & 0.6974 \\
  $\text{NLL}_{\text{sum}}$ & \underline{0.9325} & 0.8051 & \textbf{0.8981} & $\text{NLL}_{\text{sum}}$ & \underline{0.8823} & \underline{0.7799} & \underline{0.8523} \\
  LNTP       & 0.8746 & 0.7869 & 0.8683 & LNTP       & 0.5400 & 0.7192 & 0.5127 \\
  MTP        & 0.8526 & 0.7348 & 0.7899 & MTP        & 0.7957 & 0.6758 & 0.6974 \\
  PPL        & 0.8746 & 0.7869 & 0.8683 & PPL        & 0.5400 & 0.7192 & 0.5127 \\
  \midrule
  \multicolumn{4}{c}{\textit{\textsc{Gemma-3 4B}}} & \multicolumn{4}{c}{\textit{\textsc{Gemma-3 12B}}} \\
  \midrule
  Mean       & 0.6456 & 0.7118 & 0.8223 & Mean       & 0.5522 & 0.6899 & 0.7891 \\
  Std Dev    & 0.7130 & 0.7316 & 0.8473 & Std Dev    & 0.6477 & 0.7097 & 0.8118 \\
  Max        & 0.8453 & 0.7267 & 0.8328 & Max        & 0.8334 & 0.7403 & 0.8196 \\
  Q10        & 0.8013 & 0.7354 & \underline{0.8907} & Q10        & 0.8061 & \underline{0.7608} & 0.8487 \\
  Q25        & 0.8099 & 0.7425 & \textbf{0.8916} & Q25        & 0.8322 & \textbf{0.7644} & \textbf{0.8536} \\
  Q50        & 0.7720 & 0.7275 & 0.8754 & Q50        & 0.7334 & 0.7312 & 0.8377 \\
  Q75        & 0.6560 & 0.6955 & 0.8299 & Q75        & 0.5661 & 0.6891 & 0.7907 \\
  Q90        & 0.6080 & 0.6944 & 0.8023 & Q90        & 0.5108 & 0.6656 & 0.7712 \\
  Skewness   & 0.4843 & 0.6241 & 0.7273 & Skewness   & 0.4044 & 0.5871 & 0.7008 \\
  Kurtosis   & 0.4700 & 0.6087 & 0.7153 & Kurtosis   & 0.3963 & 0.5674 & 0.6913 \\
  \midrule
  SEA        & \textbf{0.8672} & \underline{0.7760} & 0.8768 & SEA        & \textbf{0.8880} & 0.7416 & \underline{0.8495} \\
  $\text{NLL}_{\text{avg}}$ & 0.6328 & 0.7059 & 0.8101 & $\text{NLL}_{\text{avg}}$ & 0.5273 & 0.6828 & 0.7758 \\
  $\text{NLL}_{\text{max}}$ & 0.7513 & 0.6890 & 0.7246 & $\text{NLL}_{\text{max}}$ & 0.8088 & 0.7191 & 0.7379 \\
  $\text{NLL}_{\text{sum}}$ & \underline{0.8634} & \textbf{0.7807} & 0.8764 & $\text{NLL}_{\text{sum}}$ & \underline{0.8826} & 0.7377 & 0.8478 \\
  LNTP       & 0.6328 & 0.7059 & 0.8101 & LNTP       & 0.5273 & 0.6828 & 0.7758 \\
  MTP        & 0.7513 & 0.6890 & 0.7246 & MTP        & 0.8088 & 0.7191 & 0.7379 \\
  PPL        & 0.6328 & 0.7059 & 0.8101 & PPL        & 0.5273 & 0.6828 & 0.7758 \\
  \bottomrule
  \end{tabular*}

  \caption{AUROC for entropy profile summaries and UQ baselines across remaining models. Skewness, Kurtosis, LNTP, and MTP report $1-\text{AUROC}$. Accumulation and extreme values are consistently the most discriminative metrics.}
  \label{tab:entropy_analysis_appendix}
\end{table*}

\subsection{Hardware and Compute}

All experiments were run on a single NVIDIA A6000 GPU (48GB VRAM). Total inference time was approximately a week across all nine LLMs, ten benchmarks and probe training.

\section{Decoupling between slice MAE and per-instance AUROC}
\label{sec:comparison_MAE_auroc}

Good domain-level estimates do not require good instance-level
predictions. MAE captures the bias of $\hat{A}(D)$ against $A(D)$,
while per-instance AUROC measures how well $\hat{P}(x)$ separates
correct from incorrect generations. The averaging in Eq.~\ref{eq:domain_acc}
decouples the two: per-instance errors that balance across $D$
cancel under averaging but still suppress AUROC. Aggregate
estimation therefore depends on the \emph{calibration} of
$\hat{P}(x)$, not its discrimination.

\begin{figure}[h]
    \centering
    \includegraphics[width=0.9\linewidth]{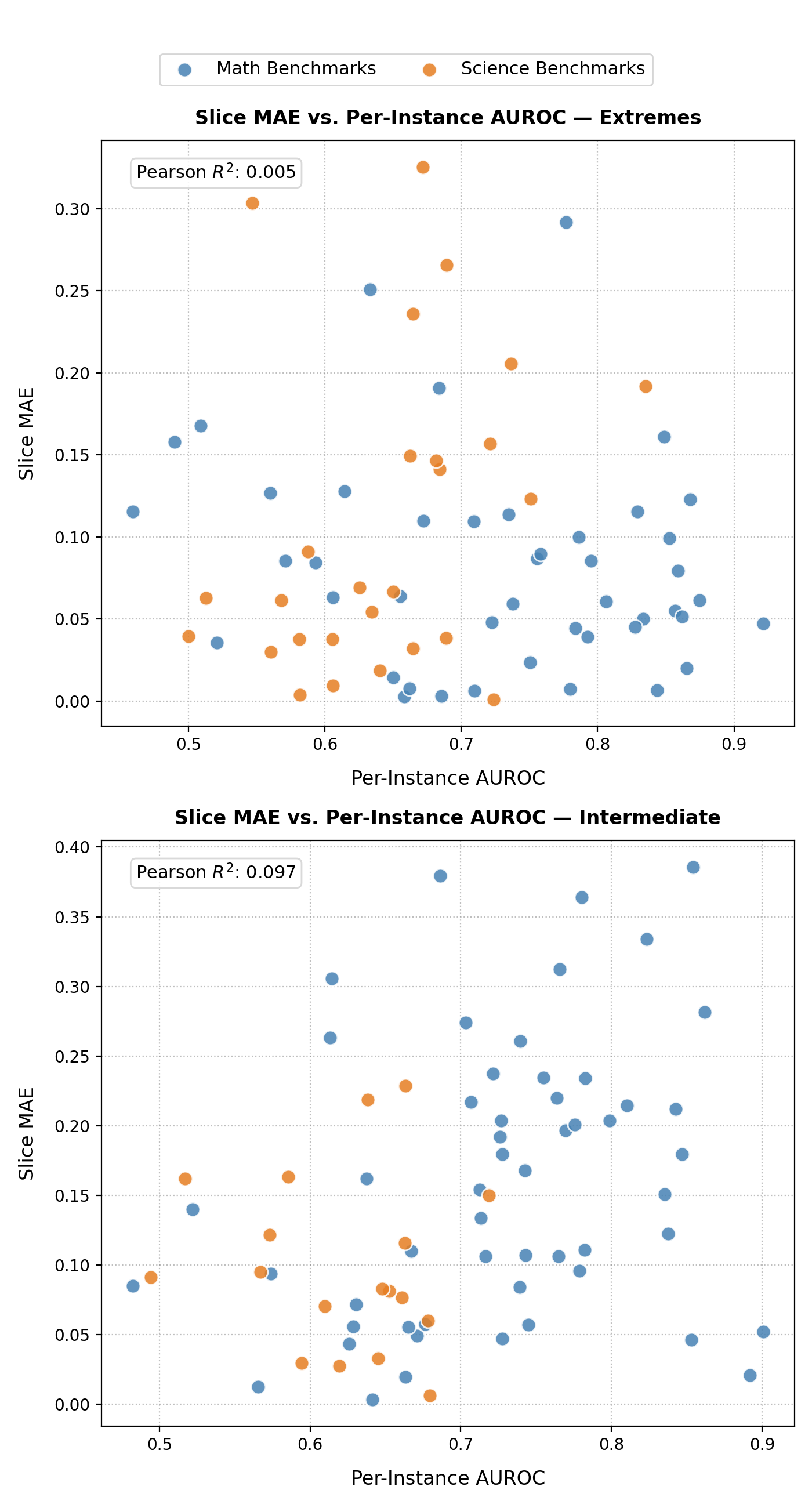}
    \caption{Slice MAE vs.\ per-instance AUROC under the Extremes and Intermediate
    estimators. Each point is one (LLM, held-out benchmark) pair.
    Pearson $R^2 = 0.005$ for Extremes and $R^2 = 0.097$ for Intermediate: slice-level error remains low across
    the full AUROC range.}
    \label{fig:MAE_auroc_ext}
\end{figure}

Figure~\ref{fig:MAE_auroc_ext} showcases this. Each point is one (LLM, held-out benchmark) pair
under the Extremes and Intermediate estimators. Pearson $R^2$ is
$0.005$ and $0.097$: slice-level error stays low across the full
AUROC range, including pairs where per-instance separation is
near chance.

\section{Additional Training Sensibility Results}
\label{sec:appendixsensibilityresults}

This appendix provides supplementary analyses of how training data composition affects accuracy estimation quality across all nine LLMs.

\paragraph{Best and Worst Benchmark Combinations.}

Table~\ref{tab:benchmark_selection} identifies the highest- and lowest-performing benchmark combinations at each $k$, along with their weighted average group accuracy. The results directly corroborate the U-shaped relationship from Figure~\ref{fig:all_llm_accuracy}: best-performing groups fall within the intermediate accuracy regime ($0.4$--$0.7$).2

For $k \geq 2$, the best combinations pair elementary benchmarks (\textsc{GSM8K}) with difficult ones (\textsc{OlympiadBench}), achieving group accuracies squarely in the optimal range. This difficulty-spanning composition exposes the estimator to both low-entropy success patterns and high-entropy failure patterns, enabling robust transfer to unseen domains. Notably, the $k=2$ combination (\textsc{GSM + Oly}) achieves median MAE of $0.087$, matching the best $k=4$ configuration despite using half the supervision. In contrast, all worst-performing combinations consist exclusively of difficult benchmarks, systematically underrepresenting low-entropy success signatures and causing the estimator to miscalibrate on easier test domains. This asymmetry reinforces a central finding: difficulty diversity in the supervision set is the main driver of better accuracy estimation.

\begin{table}[t]
\centering
\small
\begin{tabular}{clcc}
\toprule
$k$ & \textbf{Benchmarks} & \textbf{MAE} & \textbf{Acc.} \\
\midrule
\multicolumn{4}{c}{\textit{Best Performing}} \\
\midrule
1 & \textsc{Oly} & .129\textsubscript{.063} & .260 \\
2 & \textsc{GSM, Oly} & .087\textsubscript{.047} & .636 \\
3 & \textsc{GPQA, GSM, Oly} & .087\textsubscript{.046} & .531 \\
4 & \textsc{GPQA, GSM, Mat, Oly} & .086\textsubscript{.048} & .413 \\
\midrule
\multicolumn{4}{c}{\textit{Worst Performing}} \\
\midrule
1 & \textsc{Live} & .269\textsubscript{.119} & .114 \\
2 & \textsc{GPQA, Live} & .259\textsubscript{.093} & .143 \\
3 & \textsc{GPQA, Live, Mat} & .216\textsubscript{.072} & .197 \\
4 & \textsc{GPQA, Live, Mat, Sci} & .191\textsubscript{.064} & .220 \\
\bottomrule
\end{tabular}
\caption{Best and worst benchmark combinations for each k value. Median MAE with IQR shown as subscripts. Acc.\ reports weighted average group accuracy across LLMs. Abbreviations: \textsc{Oly}=OlympiadBench, \textsc{GSM}=GSM8K, \textsc{Mat}=MatSciBench, \textsc{Live}=LiveMathBench, \textsc{Sci}=SciBench.}
\label{tab:benchmark_selection}
\end{table}

\section{Per Model Feature Correlation}
\label{sec:feature_monitoring_agreement}

Presented in table~\ref{tab:feature_monitoring_agreement}.

\begin{table*}[t]
  \centering
  \scriptsize
  \setlength{\tabcolsep}{3pt}
  \renewcommand{\arraystretch}{0.88}
\resizebox{\textwidth}{!}{%
\begin{tabular}{lrrrrrrrrrrrrrrrrr}
\toprule
Model & mean & std & max & q10 & q25 & q50 & q75 & q90 & skew. & kurt. & se\_sum & nll\_avg & nll\_max & nll\_sum & lntp & mtp & ppl \\
\midrule
gemma3-12b    & 0.936 & 0.941 & 0.607 & 0.399 & 0.638 & 0.886 & \textbf{0.945} & 0.944 & -0.927 & -0.868 & 0.814 & 0.936 & 0.711 & 0.819 & 0.390 & 0.723 & 0.931 \\
gemma3-4b     & 0.910 & 0.915 & 0.817 & 0.515 & 0.651 & 0.874 & 0.911 & \textbf{0.915} & -0.890 & -0.877 & 0.817 & 0.910 & 0.790 & 0.827 & 0.532 & 0.774 & 0.906 \\
phi3-3b       & 0.910 & 0.894 & 0.827 & 0.364 & 0.537 & 0.891 & \textbf{0.911} & 0.903 & -0.878 & -0.835 & 0.140 & 0.901 & 0.780 & 0.204 & 0.479 & 0.769 & 0.899 \\
llama3-8b     & 0.776 & 0.812 & -0.136 & 0.090 & 0.383 & 0.712 & 0.776 & 0.798 & \textbf{-0.844} & -0.539 & 0.030 & 0.759 & -0.300 & 0.083 & 0.582 & -0.268 & 0.743 \\
\midrule
ministral3-8b & 0.599 & \textbf{0.645} & 0.502 & 0.285 & 0.403 & 0.571 & 0.624 & 0.629 & -0.592 & -0.577 & 0.553 & 0.608 & 0.537 & 0.573 & 0.603 & 0.531 & 0.596 \\
qwen3-4b      & 0.501 & 0.495 & 0.354 & 0.143 & 0.370 & 0.487 & 0.493 & 0.467 & -0.165 & 0.145 & 0.464 & 0.483 & 0.471 & 0.456 & \textbf{0.507} & 0.496 & 0.491 \\
qwen3-8b      & 0.371 & 0.439 & -0.027 & 0.192 & 0.427 & 0.401 & 0.309 & 0.418 & -0.274 & 0.010 & -0.414 & 0.288 & \textbf{-0.665} & -0.458 & 0.067 & -0.651 & 0.289 \\
ministral3-3b & 0.306 & -0.082 & -0.298 & 0.252 & 0.283 & 0.304 & 0.307 & 0.306 & \textbf{-0.345} & -0.133 & 0.168 & 0.343 & 0.080 & 0.249 & 0.337 & 0.064 & 0.324 \\
oss-20b       & -0.434 & -0.435 & 0.106 & \textbf{-0.556} & -0.486 & -0.418 & -0.409 & -0.482 & 0.427 & 0.439 & -0.176 & -0.416 & -0.389 & -0.223 & -0.485 & -0.411 & -0.411 \\
\bottomrule
\end{tabular}}
\caption{Per-model correlations for each feature. Largest absolute correlation per row in bold.}
\label{tab:feature_monitoring_agreement}
\end{table*}

\section{License}

Code for the paper is released under the MIT License.

\section{Use of AI-Assitants}

AI assistants were used for spell checking and proofreading of the manuscript, and for assistance with minor utility scripts. No AI-generated text was included directly in the paper, and all research, methodology, and analysis were conducted by the authors.

\end{document}